\documentclass[lettersize,journal]{IEEEtran}
\usepackage{amsmath,amsfonts}
\usepackage{algorithmic}
\usepackage{array}
\usepackage[caption=false,font=normalsize,labelfont=sf,textfont=sf]{subfig}
\usepackage{textcomp}
\usepackage{stfloats}
\usepackage{url}
\usepackage{verbatim}
\usepackage{graphicx}
\def\BibTeX{{\rm B\kern-.05em{\sc i\kern-.025em b}\kern-.08em
    T\kern-.1667em\lower.7ex\hbox{E}\kern-.125emX}}
\usepackage{balance}
\usepackage{orcidlink}
\usepackage{float}

\usepackage[utf8]{inputenc} 
\usepackage[T1]{fontenc}    
\usepackage{hyperref}       
\usepackage{url}            
\usepackage{booktabs}       
\usepackage{amsfonts}       
\usepackage[ruled,vlined]{algorithm2e} 
\usepackage{graphicx}       
\usepackage{nicefrac}       
\usepackage{microtype}      
\usepackage{xcolor}         
\usepackage{multirow}
\usepackage{amssymb}
\usepackage{amsmath}

\usepackage{wrapfig} 
\usepackage{booktabs}  
\usepackage{lipsum}    

\begin{document}
\title{Causal Debiasing Medical Multimodal Representation Learning with Missing Modalities}

\author{
    Xiaoguang Zhu\orcidlink{0000-0001-9554-2133},~\IEEEmembership{Member,~IEEE}, 
    Lianlong Sun\orcidlink{0009-0005-3208-9075}, 
    Yang Liu\orcidlink{0000-0002-1312-0146},~\IEEEmembership{Member,~IEEE},
    Pengyi Jiang\orcidlink{0009-0001-7413-9078},~\IEEEmembership{Student Member,~IEEE},
    Uma Srivatsa\orcidlink{0000-0003-0378-2346},
    Nipavan Chiamvimonvat\orcidlink{0000-0001-9499-8817},
    Vladimir Filkov\orcidlink{0000-0003-0492-4393}
\thanks{This work was supported by National Institutes of Health under Grant NIH R01 HL170520-01. Corresponding author: Xiaoguang Zhu.}
\thanks{Xiaoguang Zhu is with the DataLab: Data Science and Informatics, University of California, Davis, California 95616, USA (email: xgzhu@ucdavis.edu).}
\thanks{Lianlong Sun is with the Department of Electrical and Computer Engineering, University of Rochester, New York 14620, USA  (email: lianlongsun@rochester.edu).}
\thanks{Yang Liu is with the Academy for Engineering \& Technology, Fudan University, Shanghai 200433, China, and also with the Department of Computer Science, University of Toronto, Ontario, M5S 1A1, Canada (email: yangliu@cs.toronto.edu).}
\thanks{Pengyi Jiang is with the Department of Electrical and Computer Engineering, New York University, New York 10012, USA (email: pj2366@nyu.edu).}
\thanks{Uma Srivatsa is with UC Davis Health, California 95817, USA (email: unsrivatsa@health.ucdavis.edu).}
\thanks{Nipavan Chiamvimonvat is with the Department of Basic Medical Sciences, University of Arizona, Arizona 85004, USA (email: nchiamvimonvat@arizona.edu).}
\thanks{Vladimir Filkov is with the Department of Computer Science, University of California, Davis, California 95616, USA (email: vfilkov@ucdavis.edu).}
}

\markboth{IEEE Transactions on Knowledge and Data Engineering,~Vol.~XX, No.~X, AUGUST~2025}%
{Causal Debiasing Medical Multimodal Representation Learning with Missing Modalities}

\maketitle

\begin{abstract}
Medical multimodal representation learning aims to integrate heterogeneous clinical data into unified patient representations to support predictive modeling, which remains an essential yet challenging task in the medical data mining community. However, real-world medical datasets often suffer from missing modalities due to cost, protocol, or patient-specific constraints. Existing methods primarily address this issue by learning from the available observations in either the raw data space or feature space, but typically neglect the underlying bias introduced by the data acquisition process itself. In this work, we identify two types of biases that hinder model generalization: \textit{missingness bias}, which results from non-random patterns in modality availability, and \textit{distribution bias}, which arises from latent confounders that influence both observed features and outcomes. To address these challenges, we perform a structural causal analysis of the data-generating process and propose a unified framework that is compatible with existing direct prediction-based multimodal learning methods. Our method consists of two key components: (1) a missingness deconfounding module that approximates causal intervention based on backdoor adjustment and (2) a dual-branch neural network that explicitly disentangles causal features from spurious correlations. We evaluated our method in real-world public and in-hospital datasets, demonstrating its effectiveness and causal insights.
\end{abstract}


\section{Introduction}
Medical data is inherently multimodal, consisting of a wide range of heterogeneous modalities such as electronic health records (EHR) \cite{johnson2023mimic}, electrocardiograms (ECG) \cite{wagner2020ptb}, biomarkers \cite{maltesen2020longitudinal}, and medical imaging \cite{allen2014uk}, making medical data mining significantly more challenging compared to single-modal knowledge discovery tasks. To address this complexity in healthcare data engineering, researchers have proposed multimodal representation learning approaches that aim to combine information between heterogeneous data sources to learn unified latent representations capturing complementary and correlated patterns, which facilitate downstream data mining applications such as disease classification, progression forecasting, and prediction of treatment outcome \cite{kline2022multimodal}. However, clinical data are often incomplete in practice, where specific modalities are missing for certain patients due to various practical constraints \cite{huang2020fusion}. This prevalent data incompleteness poses significant challenges for conventional multimodal learning models, which typically assume that all modalities are fully observed during both the training and the testing phases \cite{ramachandram2017deep}.

To address this challenge, some earlier work \cite{wang2020multimodal,ni2019modeling} disregarded incomplete data in the training phase, and therefore were limited in their real-world applicability, where data are scarce and missingness is unavoidable. Current methods fall predominantly into two categories when addressing the missing modalities problem: (1) modality imputation and (2) direct prediction approaches. The former \cite{ngiam2011multimodal,ma2021smil,tran2017missing} use deep generative models to reconstruct the absent modalities. However, these methods rely on strong assumptions to learn the mappings from a lower-dimensional latent space to the high-dimensional original input space \cite{wu2024multimodal}, which is an ill-posed inverse problem \cite{kabanikhin2008definitions}. The latter, direct prediction approaches, directly predict outcomes from incomplete observed inputs. Recent methods adopt Graph Neural Networks (GNNs) \cite{chen2020hgmf,zhang2022m3care,wu2024multimodal} or Transformer \cite{ma2022multimodal,kim2021vilt} to model inter-modality or inter-patient correlations to compensate for missing information in the feature space.

Despite the progress of both approaches, they had a common limitation that they overlook the correlations between the modalities, neglecting the bias introduced by the data acquisition process \cite{li2022more}. In this way, the model could be biased towards modality-specific patterns that are overrepresented in the observed data. This issue is further exacerbated by the fact that medical data often contains features that are not causally related to the underlying disease mechanisms, making models more likely to exploit spurious shortcuts based on non-causal correlations \cite{castro2020causality}. In this paper, we identify the following two types of biases: \textbf{i) Distribution Bias.} Clinical data often contains spurious correlations introduced by latent confounders affecting both observed data and outcomes, with missingness further amplifying these confounding effects \cite{cornelisz2020addressing}. For example, in Alzheimer’s research, visible brain atrophy on MRI is more common in late stage patients. This may cause models to over-rely on imaging features like ventricular enlargement, treating them as direct predictors of cognitive decline, even when such features may result from aging or comorbidity unrelated to AD progression. Such correlations lead the model to learn nongeneralizable patterns related to the data collection distribution rather than true disease mechanisms, limiting the generalization ability across hospitals or patient populations. \textbf{ii) Missingness Bias.} Prior approaches often assume that missingness is random \cite{zhang2022m3care,you2020handling}, without modeling the underlying mechanisms behind why certain modalities are missing. In reality, missing data can be the result of factors such as high cost \cite{ford2000non}, patient dropout \cite{pan2021disease}, noncompliance \cite{ramos2004mri}, or technical failures \cite{chen2020hgmf}. For example, neuroimaging data in ADNI \cite{jack2008alzheimer} are frequently missing for patients with advanced dementia due to enrollment restrictions. Such structured, non-random missingness reflects latent patient conditions or institutional policies, making the absence of modalities itself a potential source of bias in downstream prediction.

\begin{figure*}
    \centering
    \includegraphics[width=0.99\linewidth]{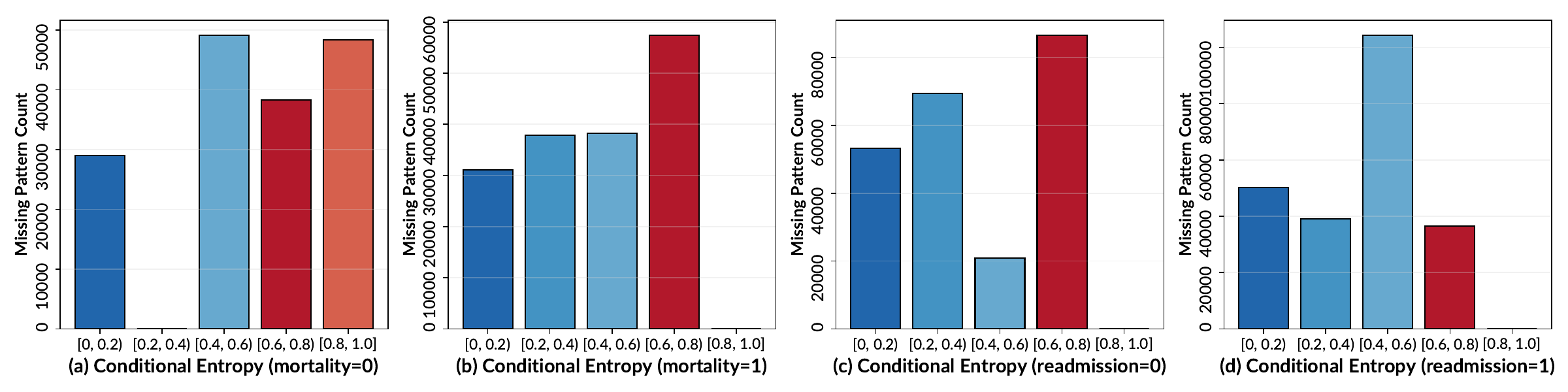}
    \caption{Conditional entropy analysis on the MIMIV-IV \cite{johnson2023mimic} dataset for mortality prediction. More samples with normalized zero-conditional entropy reveal a strong presence of the bias.}
    \label{fig:entropy}
\end{figure*}

We have carried out a toy experiment to verify missingness bias in the MIMIC-IV dataset \cite{johnson2023mimic}, which aims to quantify the extent to which outcome labels are correlated with missingness patterns. For each sample, we construct a 6-dimensional binary missingness vector based on the availability of the following modalities: demographics, diagnosis codes, procedure codes, medication records, lab values, and clinical notes. In the mortality task, we observe 47 different missingness patterns, while the readmission task contains 54 patterns. We compute the normalized conditional entropy (NCE) \cite{panda2018contemplating} of the label given each missingness pattern. A lower NCE indicates that the outcome is more predictable given the pattern, i.e., the pattern exhibits stronger label-specific bias. We visualize the distribution of these patterns in terms of the total number of samples associated with each range for both mortality (Fig.~\ref{fig:entropy}(a, b) and readmission tasks (Fig.~\ref{fig:entropy}(c, d). Concretely, approximately only 29\% of the samples with mortality = 0 fall into missingness patterns whose NCE is in the range [0, 0.2), indicating that these patterns are highly predictive of negative outcomes. These observations suggest that both outcome groups exhibit substantial pattern-label correlation, reflecting a strong dependency between missingness configurations and mortality outcomes, and highlighting the potential biases in the data.

To overcome biases, we attempt to incorporate causal inference \cite{pearl2009causality} into the medical multimodal representation learning task. In this paper, we introduce a unified \textbf{Ca}usal \textbf{D}ebiasing (\textbf{CaD}) framework that systematically identifies and mitigates both missingness and distributional biases through structural causal modeling. In detail, CaD consists of two key modules: (1) A \textit{Missingness Deconfounding Module} that approximates interventional distributions via a Normalized Weighted Geometric Mean, leveraging a learned confounder dictionary to perform efficient backdoor adjustment over missingness patterns. (2) A \textit{Causal-Biased Disentanglement Module} that uses a dual-branch neural network to separate patient representations into invariant causal features and spurious biased features, guided by mutual information minimization and counterfactual learning. Importantly, our framework is model-agnostic and can be flexibly integrated into existing direct prediction approaches with minimal slight modification. We validate our approach on three public real-world medical datasets (i.e., MIMIC-IV \cite{johnson2023mimic}, eICU \cite{pollard2018eicu} and ADNI \cite{jack2008alzheimer}) and one in-hospital dataset (i.e., AFib~\footnote{Unpublished, collected by two of the authors under IRB-approved protocol.}), demonstrating consistent gains in both predictive performance and robustness under missingness.

The contributions of this work are summarized as follows:
\begin{itemize}
    \item We provide the first systematic causal analysis of bias sources in medical multimodal learning with missing modalities. Through structural causal modeling, we identify and formalize two distinct types of biases: \emph{missingness bias} arising from non-random modality availability, and \emph{distribution bias} caused by latent confounders affecting both observed features and outcomes.
    \item  We propose CaD, a model-agnostic framework that systematically mitigates both identified biases through principled causal inference. The framework consists of MDM that approximates interventional distributions via backdoor adjustment and CBDM that separates causal from spurious features using dual-branch neural networks with counterfactual learning.
    \item We conducted extensive experiments on four real-world medical datasets spanning different clinical domains. The results show that CaD consistently improves predictive performance in multiple baseline methods, with AUC-PRC improvements of up to 3.19\% in mortality prediction and 5.12\% in readmission tasks.
\end{itemize}

The remainder of this paper is organized as follows. Sec.~\ref{sec2} reviews related work on medical multimodal learning and causal debiasing approaches. Sec.~\ref{sec3} presents our problem formulation and a detailed causal analysis of bias sources using structural causal models. Sec.~\ref{sec4} describes the proposed CaD framework, including the design of both debiasing modules and their integration strategy. Sec.~\ref{sec5} provides experimental evaluation on four datasets, including comparisons with state-of-the-art baselines, ablation studies, and robustness analysis. Finally, Sec.~\ref{sec7} concludes the paper. 

\section{Related works}~\label{sec2}
\subsection{Medical Multimodal Learning with Missing Modalities}
Multimodal learning has gained substantial traction in healthcare, leveraging heterogeneous data such as clinical notes, imaging, tabular labs, and waveforms to improve diagnosis and prognosis tasks \cite{ramachandram2017deep}. Models like RAIM \cite{xu2018raim} and MedFuse \cite{hayat2022medfuse} propose architectures to jointly encode continuous and discrete modalities for monitoring patient status. However, a major challenge arises from the modality level missingness, which is prevalent in clinical datasets due to cost, feasibility, or clinical decision pathways. The previous imputation-based methods directly reconstruct the missing modality with generative models. CM-AE \cite{ngiam2011multimodal} uses the cross-modality auto-encoder, while SMIL \cite{ma2021smil} adopts a Beyesian meta-learning approach  to recover the missing modality. However, such methods would introduce extra noise, especially when the size of samples is limited. The following direct prediction approaches leverage transformer or graph neural networks to model the relationships between modalities to impute the missing information in the feature space. MT \cite{ma2022multimodal}, UMSE \cite{lee2023learning}, and ViLT \cite{kim2021vilt} represent the missing modalities as prompts and use cross-modal attention in the Transformer to fuse the embeddings. Recently, graph-based approaches have been proposed. GRAPE \cite{you2020handling} and MUSE \cite{wu2024multimodal} use a bipartite graph to model the features of the modality as edges, improving the feasibility of representing the missing modalities. HGMF \cite{zhang2022m3care} leverages the hypergraph to reason the intra- and interhypernode interactions. M3Care \cite{zhang2022m3care} imputes the task-related information of missing modalities in the latent space by the auxiliary information from the similar neighbors of each patient with GNN. However, existing methods generally assume missingness-at-random and rarely account for the confounding factors underlying missing patterns. Our work bridges this gap by explicitly modeling missingness from a causal perspective and can be extended upon the previous baselines.

\subsection{Causal Debiasing in Representation Learning}
Causal representation learning \cite{scholkopf2021toward} seeks to isolate invariant causally relevant characteristics while removing spurious correlations induced by observational biases. A growing body of work formulates this through structural causal models \cite{pearl2009causality}, which provide a principled framework for modeling interventions and counterfactual reasoning. By explicitly accounting for the causal structure underlying the data, causal debiasing methods can improve the generalizability of the model. These approaches have shown promising results in a variety of computer vision tasks \cite{shao2024knowledge, li2024multimodal, zhou2025learning}, such as video anomaly detection \cite{liu2025crcl}, visual recognition \cite{liu2022contextual}, recommendation \cite{wang2022causal,wei2022causal}, and scene graph generation \cite{liu2025causal}. In graph-based learning, disentanglement-based methods such as DisC \cite{fan2022disentangled} and CausalVAE \cite{yang2021causalvae} aim to learn representations where causal and bias-related features are separated, improving generalization under distribution shift. Divyansh \textit{et al.} \cite{kaushik2020counterfactual} pursue domain-agnostic representations by proposing counterfactual data augmentation. Although Shachi \textit{et al.} \cite{deshpande2022deep} use SCM to estimate the causal effect for incomplete multimodal data, they assume the availability of complete training data. Despite promising progress, most causal debiasing efforts have focused on unimodal or graph-structured data, and extending these ideas to complex multimodal healthcare settings with missing modalities remains underexplored. Our work extends these ideas by proposing a causal framework that addresses both modality missingness and distribution biases simultaneously in a medical multimodal setting.

\section{Preliminaries and Problem Analysis}~\label{sec3}
\label{fullanalysis}
\subsection{Preliminaries}
In medical multimodal datasets, we denote the multimodal data for patient \( p \) as \( \mathbf{X}^{(p)} = (\mathbf{x}_1^{(p)}, \mathbf{x}_2^{(p)}, \dots, \mathbf{x}_M^{(p)}) \), where \( M \) is the number of modalities. Correspondingly, \( \mathbf{y}^{(p)} \) represents the label associated with predictive tasks. We represent the modality missingness pattern for patient \( p \) as a binary vector \( \mathbf{M}^{(p)} \in \{0,1\}^M \), where \( \mathbf{M}^{(p)}[m] = 1 \) indicates the availability of modality \( m \) and \( 0 \) otherwise. Consequently, the observed data \( \mathbf{X}_O^{(p)} \) is derived by masking \( \mathbf{X}^{(p)} \) with \( \mathbf{M}^{(p)} \). Given a data set \( \mathcal{D}_{tr} = \{ (\mathbf{X}_O^{(p)}, \mathbf{M}^{(p)}, \mathbf{y}^{(p)}) \}_{p=1}^N \) comprising \( N \) patients with partially observed modalities, our goal is to learn a multimodal model \( f_{\Theta}(\cdot) \) that robustly predicts clinical outcome \( \mathbf{y}^{(p)} \) despite the challenges introduced by the lack of modality and latent confounder. Formally, conventional multimodal learning optimizes the following:
\begin{equation}
    \scalebox{0.95}{$\displaystyle
\arg\min_{\Theta} \mathbb{E}_{(\mathbf{X}_O^{(p)}, \mathbf{M}^{(p)}, \mathbf{y}^{(p)}) \sim \mathcal{D}_{tr}} \left[ \mathcal{L}(f_{\Theta}(\mathbf{X}_O^{(p)}, \mathbf{M}^{(p)}), \mathbf{y}^{(p)}) \right],
$}
\label{optimizationgoal}
\end{equation}
where \( \mathcal{L}(\cdot) \) denotes the task-specific loss (e.g., cross-entropy for classification).

\subsection{Problem Analysis}
\label{analysis}

\begin{figure}[htbp]
    \centering
    \includegraphics[width=0.95\linewidth]{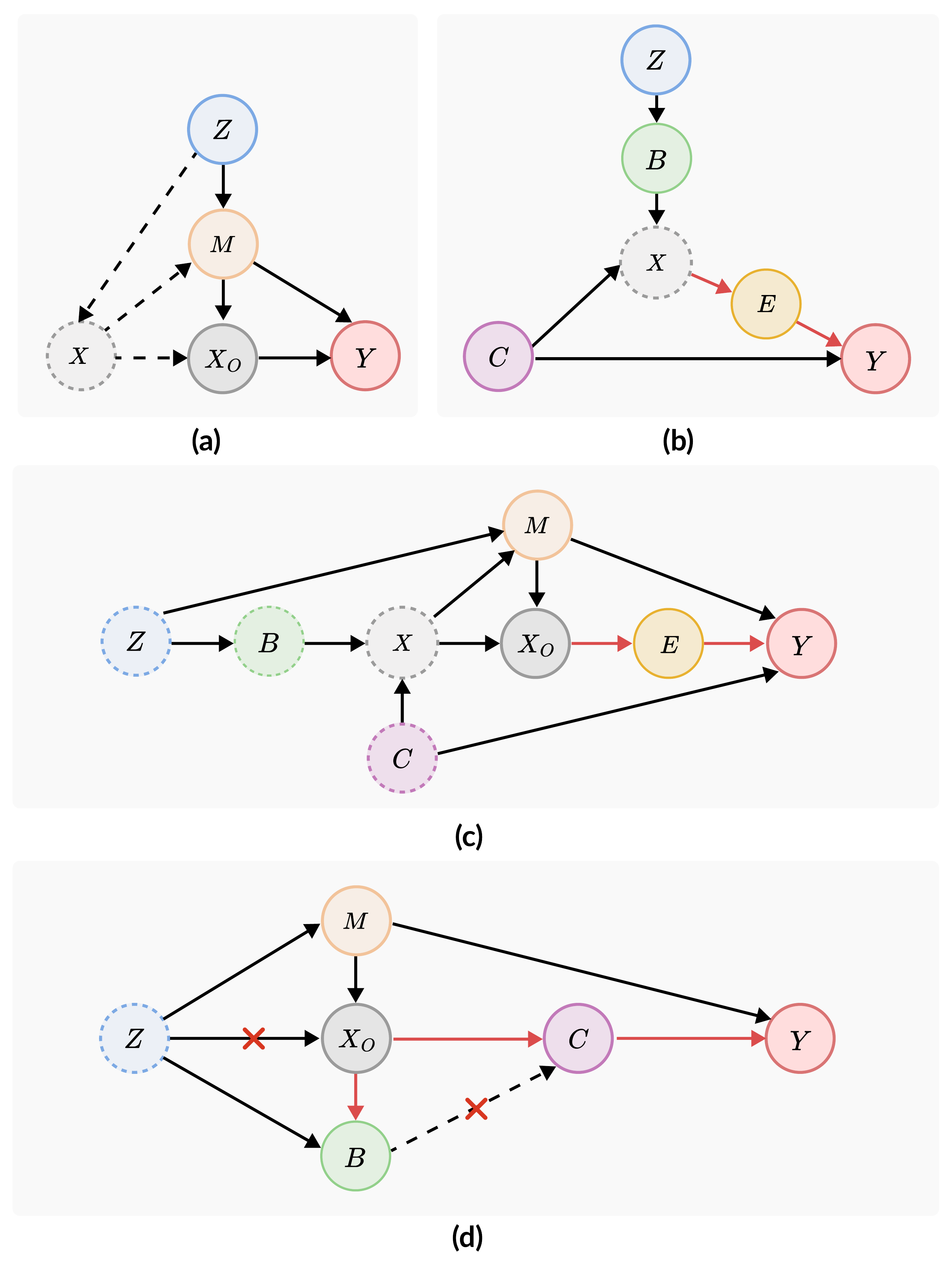}
    \caption{Illustration of the proposed structured causal models, including (a) SCM under missingness bias, (b) SCM under distribution bias, (c) Unified SCM in generation view, and (d) Proposed debiasing unified SCM in prediction view. The red lines denote the prediction view of current methods.}
    \label{fig:causalgraph}
\end{figure}

In this paper, we identify two types of bias introduced by the latent confounder. Therefore, we formalize the data generation process and the behavior of the model using the structured causal model (SCM) \cite{pearl2016causal,pearl2009causality} that consists of the following variables: unobservable multimodal data $X$, observable multimodal data with missingnes $\mathbf{X}_O$, modality mask $\mathbf{M}$, latent confounder $\mathbf{Z}$, causal variable $\mathbf{C}$, bias variable $\mathbf{B}$, and ground truth label/prediction $\mathbf{Y}$. We illustrate the causal graphs of \textit{SCM under Missingness Bias} and \textit{SCM under Distribution Bias} in Fig.~\ref{fig:causalgraph}(a) and (b), respectively. 

\subsubsection{SCM under Missingness Bias}
Firstly, we use the SCM shown in Fig.~\ref{fig:causalgraph}(a) to illustrate the missingness bias, where the detailed causal relationships are described below.

$\bullet$ $\mathbf{Z} \rightarrow \mathbf{M} \rightarrow \mathbf{X}_O$. This link captures how the latent confounder $Z$ (e.g., acquisition bias, demographics) influences the availability of observable modalities via the missingness mask $\mathbf{M}$. For example, certain populations might be systematically missing specific modality data \cite{RAJENDRAN2024100913,colwell2022patient} (e.g., MRI), directly affecting the observed data $\mathbf{X}_O$. 

$\bullet$ {$\mathbf{X} \rightarrow \mathbf{X}_{O}$, $\mathbf{X}\rightarrow  \mathbf{M} \rightarrow \mathbf{X}_{O}$.} The dashed links represent the observed multimodal data generated by the unobserved complete data $\mathbf{X}$ and the missing mask $\mathbf{M}$. These links highlight that the observed data is a selection-biased subset influenced by unobserved mechanisms.

\textit{Remark 1. The causal mechanism from $\mathbf{X}$ to the missingness indicator $\mathbf{M}$ characterizes the nature of the missingness: If $\mathbf{M}$ is independent of $\mathbf{X}$, the missingness is \emph{Missing Completely At Random (MCAR)}  . If $\mathbf{M}$ depends only on the observed data $\mathbf{X}_O$, the mechanism is \emph{Missing At Random (MAR)}. If $\mathbf{M}$ depends on the unobserved values in $\mathbf{X}$, the mechanism is \emph{Missing Not At Random (MNAR)}}.

$\bullet$ $\mathbf{Z} \rightarrow \mathbf{M} \rightarrow \mathbf{Y}$, $\mathbf{X}_O \rightarrow \mathbf{Y}$. The observed modalities $\mathbf{X}_O $ are used to predict the label $\mathbf{Y}$. However, since $\mathbf{X}_O $ is subject to selection via $\mathbf{M}$, its predictive utility may be biased if confounding is not addressed. The confounder $\mathbf{Z}$ also directly affects the outcome $\mathbf{Y}$ via the combination of observed modalities controlled by $\mathbf{M}$, representing structural bias such as demographic factors that correlate with the prevalence of the disease or the distribution of the outcomes independently of the observed data \cite{jones2024causal}.

\subsubsection{SCM under Distribution Bias} Secondly, we introduce the SCM under distribution bias, as illustrated in Fig.~\ref{fig:causalgraph}(b). In practice, the influence of the confounder $\mathbf{Z}$ on $\mathbf{X}$ is often heterogeneous: part of the information in $\mathbf{X}$ may reflect genuine causal factors relevant to the target $\mathbf{Y}$, while the rest may encode spurious variations induced by $\mathbf{Z}$. To make this distinction explicit, we further refine the SCM by decomposing the multimodal data $\mathbf{X}$ into two latent variables $\mathbf{C}$ and $\mathbf{B}$, representing the causal features and bias features, respectively. Therefore, the SCM can be viewed as the detailed decomposition of $\mathbf{Z} \rightarrow \mathbf{X}$ in Fig.~\ref{fig:causalgraph}(a).

$\bullet$ $\mathbf{C} \rightarrow \mathbf{X} \leftarrow \mathbf{B}$. This link represents that the complete data is generated by causal features $\mathbf{C}$ and bias features $\mathbf{B}$, which are both unobserved latent variables. 

$\bullet$ $ \mathbf{Z} \rightarrow \mathbf{B}.$ This link means that the latent confounder could affect the bias features. For example, age is a well-known confounder in Alzheimer's disease \cite{pichet2023confounding}, which correlates with changes in brain structure and disease diagnosis, but is not necessarily a causal feature of Alzheimer's progression. The bias features $\mathbf{B}$ could be enlarged ventricles or general gray matter that changes with normal aging. These features may correlate with a positive Alzheimer's label, but are not specific to Alzheimer's pathology, as they also occur in healthy aging individuals.

$\bullet$ $\mathbf{C} \rightarrow \mathbf{Y}$. This link shows that the causal features contribute to the prediction $\mathbf{Y}$. In our design, we explore the method that only makes predictions with endogenous causal features. 

$\bullet$ $\mathbf{X} \rightarrow \mathbf{E} \rightarrow \mathbf{Y}$. Existing methods usually learn the embedding $\mathbf{E}$ based on multimodal data $\mathbf{X}$ to make the prediction $\mathbf{Y}$, thus inevitably introducing the biased feature $\mathbf{V}$ into the final prediction.

\subsubsection{Unified SCMs}
According to the previous analysis of SCMs under missingness bias and distribution bias, we unify the two SCMs in data generation in Fig. \ref{fig:causalgraph}(c). We can clearly observe that the proportion of biased features $\mathbf{B}$ and causal features $\mathbf{C}$ can be affected by the missingness mask $\mathbf{M}$ when the missingness mechanism depends on these features, i.e., under MNAR conditions, conditioned on the observed multimodal data $\mathbf{X}_O$. Since $\mathbf{X}_O$ is obtained by applying $\mathbf{M}$ to $\mathbf{X}$, which is generated from both $\mathbf{B}$ and $\mathbf{C}$, an increase in the missing rate can selectively filter out samples in a non-uniform manner. This alters the conditional distribution $P(\mathbf{B}, \mathbf{C} \mid \mathbf{X}_O)$, leading to shifts in the relative representation of $\mathbf{B}$ and $\mathbf{C}$ in the dataset and introducing biases into the prediction task. 

However, explicitly disentangling causal and bias features is infeasible in this study. On the one hand, modality missingness hinders reliable extraction of both types of features, i.e, the complete data $\mathbf{X}$ is unobserved. On the other hand, identifying causal features requires substantial domain expertise, which is often unavailable in practice. To address these challenges, we adopt a representation-based approach: we learn disentangled embeddings of their respective effects in the latent space conditioned on the input $\mathbf{X}_O$. To achieve this, we propose a unified causal framework in the prediction view that integrates the two SCMs, as illustrated in Fig.~\ref{fig:causalgraph}(d).In the prediction process, $\mathbf{C}$ and $\mathbf{B}$ denote the disentangled embeddings for causal and biased features which is learned from the observed data $\mathbf{X}_{O}$.

In the unified SCM, the causal relationships above remain the same. We aim to predict the label $\mathbf{Y}$ only using the causal features $\mathbf{C}$, i.e., $P(\mathbf{Y}|\mathbf{C})$, where $\mathbf{C}$ is learned from the observed modalities $\mathbf{X}_O$ and is assumed to be identical to the causal feature of the complete data $\mathbf{X}$. However, according to the $d$-connection theory \cite{pearl2009causality}, we can find two paths that induce spurious correlations: (1) $\mathbf{C} \leftarrow \mathbf{X}_O\leftarrow \mathbf{M} \leftarrow \mathbf{Z} \rightarrow \mathbf{Y}$ and (2) $\mathbf{Z} \rightarrow \mathbf{B} \dashrightarrow \mathbf{C}  \rightarrow \mathbf{Y}$. These two active paths imply that both missingness bias and distribution bias can confound the prediction of $\mathbf{Y}$, as the learned representation may inadvertently capture spurious relations mediated by 
$\mathbf{M}$ and $\mathbf{B}$, rather than purely reflecting the true causal factors.

\subsection{Causal Debiasing}
The existing methods aim to solve the problem defined in Eq.~\ref{optimizationgoal}, where the likelihood is $P(\mathbf{Y}|\mathbf{M},\mathbf{X}_O)$. As illustrated in Fig.~\ref{fig:causalgraph}, this process can be formulated by Bayesian rules as:
\begin{equation}
    \scalebox{0.9}{$\displaystyle
\begin{aligned}
P(\mathbf{Y} \mid \mathbf{M}, \mathbf{X}_O) 
= \sum_{\mathbf{z}} P(\mathbf{Y} \mid \mathbf{X}_O, \mathbf{C}, \mathbf{B}) 
    \cdot P(\mathbf{z} \mid \mathbf{M}, \mathbf{X}_O),
\end{aligned}
$}
\end{equation}
where $\mathbf{C} = f_c(\mathbf{X}_O, \mathbf{z})$ and $\mathbf{B} = f_b(\mathbf{X}_O, \mathbf{z})$ denote the causal and biased features extracted from the observed modality $\mathbf{X}_O$ using two models $f_c(\cdot)$ and $f_b(\cdot)$. We can clearly observe that the latent confounder $\mathbf{Z}$ influences the prediction in two distinct ways as described in Sec.~\ref{analysis}.

\subsubsection{Missingness Bias and Backdoor Adjustment}

To address the confounding effect introduced by $\mathbf{Z}$ via the missingness mechanism $\mathbf{M}$, an intuitive solution is to intervene on $\mathbf{X}$ and ensure that each modality is fairly represented in all patients. This idea is analogous to conducting a randomized controlled experiment in which every subject is observed in all possible combinations of modality. However, such data collection is infeasible in clinical settings. Instead, we adopt a backdoor adjustment strategy \cite{pearl2016causal} to approximate the interventional distribution using observational data. Specifically, we marginalize out the latent confounder $\mathbf{Z}$ conditioned on the observed mask $\mathbf{M}$, leading to the following equation:
\begin{equation}
    \scalebox{0.9}{$\displaystyle
\begin{aligned}
P(\mathbf{Y} \mid \mathbf{M}, \textit{do}(\mathbf{X}_O)) = \sum_{\mathbf{z}} P(\mathbf{Y} \mid \mathbf{X}_O, 
    \mathbf{C}, \mathbf{B}) 
    \cdot P(\mathbf{z} \mid \mathbf{M}).
\end{aligned}
$}
\label{backdoor}
\end{equation}

\subsubsection{Distribution Bias and Causal-Biased Disentanglement.}

Although backdoor adjustment removes the confounding through $\mathbf{M}$, it does not address distribution bias. To resolve this, we further disentangle $\mathbf{C}$ and $\mathbf{B}$ such that the prediction of $\mathbf{Y}$ depends only on $\mathbf{C}$, i.e., we enforce conditional independence $\mathbf{B} \perp \mathbf{Y} \mid \mathbf{C}$. Under this assumption, the above expression simplifies to:
\begin{equation}
\begin{aligned}
P(\mathbf{Y} \mid \mathbf{M}, \textit{do}(\mathbf{X}_O)) 
&= \sum_{\mathbf{z}} P(\mathbf{Y} \mid \mathbf{X}_O, \mathbf{C} ) \cdot P(\mathbf{z} \mid \mathbf{M}) \\
&\approx P(\mathbf{Y} \mid \mathbf{C}).
\end{aligned}
\label{disentanglement}
\end{equation}

This final form reflects a deconfounded prediction based solely on the causal representation $\mathbf{C}$, free of missingness bias and representation-level distribution bias.

\textit{Remark 2. The current formulation predicts $\mathbf{Y}$ by marginalizing the latent confounder $\mathbf{Z}$, where the approximation becomes exact if and only if the latent confounder $\mathbf{z}$ is conditionally independent of the outcome given the causal representation, ie, $\mathbf{z} \perp \mathbf{Y} \mid \mathbf{C}$. This requires that the causal features have sufficient information for prediction.}

\section{Methodology}~\label{sec4}
\label{methodology}

\begin{figure*}[t]
    \centering
    \includegraphics[width=1.0\linewidth]{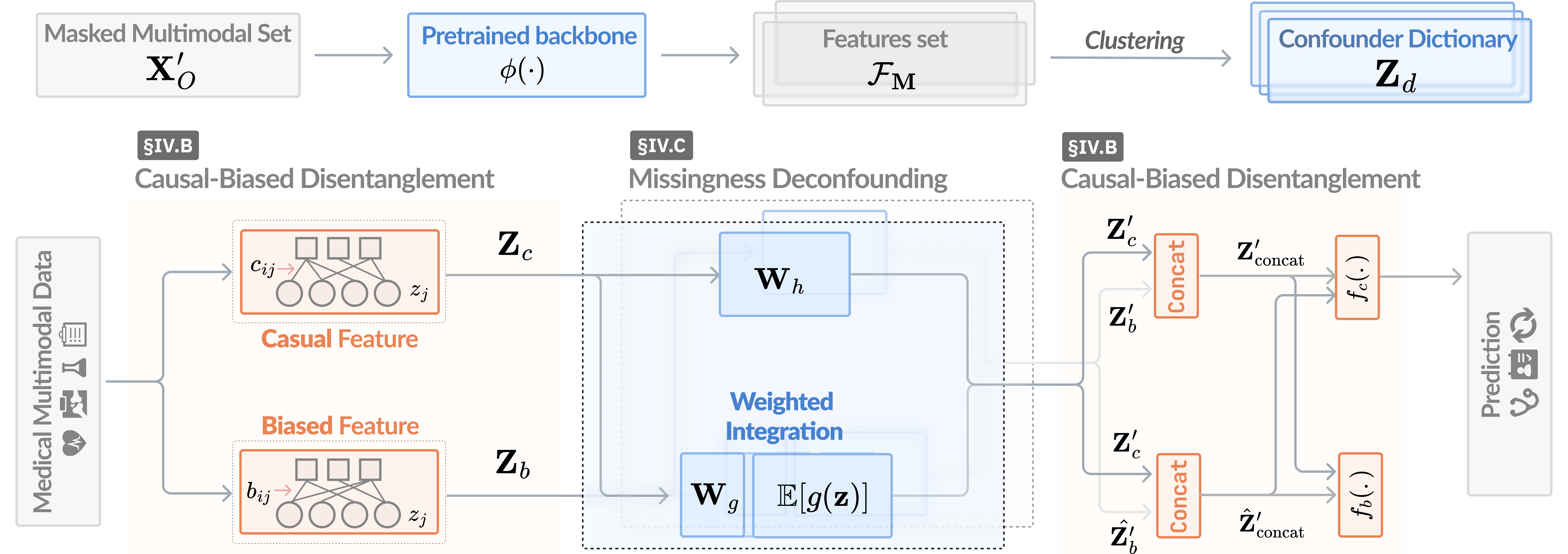}
    \caption{The overall structure of the proposed \textbf{CaD} framework. }
    \label{fig:framework}
\end{figure*}

Inspired by the causal analysis described above, we present the \textbf{Ca}usal \textbf{D}ebiasing (\textbf{CaD}) framework for medical multimodal learning with missing modalities in this section. As illustrated in Fig.~\ref{fig:framework}, the whole framework consists of two main modules: Causal-Biased Disentanglement Module (\textbf{CBDM}) and Missingness Deconfounding Module (\textbf{MDM}).  Firstly, CBDM targets the \emph{ distribution bias}, where spurious correlations arise due to the confounder’s influence on both latent causal features and biased features. Secondly, MDM addresses the \emph{missingness bias}, which arises due to the modality missingness induced by latent confounders. 

CaD is designed to be flexibly integrated into existing graph-based medical multimodal representation learning methods, such as MUSE \cite{wu2024multimodal}, M3Care \cite{zhang2022m3care}, or GRAPE \cite{you2020handling}, with light modifications to the message-passing layers and inserted after the learned feature after the original methods. The proposed CaD framework acts as a component of causal enhancement on top of the off-the-shelf direct prediction pipelines, enhancing robustness against both modality-level missingness and latent confounding. Without loss of generality, we adopt the bipartite patient-modality graph structure from GRAPE \cite{you2020handling} as the baseline for our CaD framework. 

\subsection{Bipartite Graph Neural Network}

In this setup, a batch of multimodal data is encoded as a bipartite graph $\mathcal{G} = (\mathcal{V}, \mathcal{E})$ with the node set $\mathcal{V} = \mathcal{V}_P \cup \mathcal{V}_M$ containing patient nodes $\mathcal{V}_P = \{u_1, u_2, \dots, u_N\}$ and modality nodes $\mathcal{V}_M = \{v_1, v_2, \dots, v_M\}$. An edge $\mathbf{e}_{u_p v_m} \in \mathcal{E}$ is created if modality $m$ is observed for patient $p$.

\subsubsection{Encoder Initialization}
Each edge $e_{u_p v_m}$ is initialized using the feature vector of modality $m$ for patient $p$:
\begin{equation}
\mathbf{e}_{u_p v_m}^{(0)} = \text{Encoder}_m(\mathbf{x}_m^{(p)}),
\label{eq:encoder}
\end{equation}
where $\mathbf{x}_m^{(p)}$ is the raw input feature of modality $m$ for patient $p$, and $\text{Encoder}_m$ is a modality-specific backbone encoder. For node initialization, modality nodes use one-hot encodings: $\mathbf{h}_{v_m}^{(0)} = \text{OneHot}(m)$, while patient nodes are initialized with constant vectors $\mathbf{h}_{u_p}^{(0)} = \mathbf{1}$.

\subsubsection{Message Passing}
A multilayer graph neural network (GNN) is then applied to the bipartite graph to propagate information between patients and modalities as follows:
\begin{equation}
\mathbf{Z} = \text{GNN}(\mathcal{G}).
\label{eq:graph}
\end{equation}
At each layer $l$, node $i$ updates its representation with:
\begin{equation}
    \scalebox{0.85}{$\displaystyle
\begin{aligned}
\mathbf{h}_i^{(l)} = f_A^{(l)} \Big( \mathbf{h}_i^{(l-1)}, 
\big\{ f_M^{(l)}(&\mathbf{h}_i^{(l-1)},\mathbf{h}_j^{(l-1)}, \mathbf{e}_{ji}^{(l-1)}) \mid j \in \mathcal{N}(i) \big\} \Big),
\end{aligned}
$}
\end{equation}
where $\mathcal{N}(i)$ denotes the neighbors of node $i$, and $f_A^{(l)}$ and $f_M^{(l)}$ are learnable aggregation and message functions (e.g., sum, attention, MLP). After $L$ layers of propagation, the patient node embeddings $\mathbf{Z} \in \mathbb{R}^{N \times d}$ are used for the downstream prediction. In detail, the node embeddings and edge embeddings are updated by:
\begin{equation}
    \scalebox{0.65}{$\displaystyle
\mathbf{h}_i^{(l)} = \mathbf{U}^{(l)} [\
\mathbf{h}_i^{(l-1)} \|\ 
\text{Mean} \left( \left\{ 
\text{ReLU}\left( \mathbf{W}^{(l)} \mathbf{h}_j^{(l-1)} + \mathbf{O}^{(l)} \mathbf{e}_{ji}^{(l-1)} \right) 
\mid j \in \mathcal{N}(i) \right\} \right)]
    $}
,
\label{eq:node_update}
\end{equation}

\begin{equation}
\begin{aligned}
\mathbf{h}_i^{(l)} = &\mathbf{U}^{(l)} \Big[\, \mathbf{h}_i^{(l-1)} \,\big\|\, 
\text{Mean} \big( \big\{\, 
\text{ReLU}\big(&\mathbf{W}^{(l)} \mathbf{h}_j^{(l-1)} \\
&+\, \mathbf{O}^{(l)} \mathbf{e}_{ji}^{(l-1)} \big) \mid j \in \mathcal{N}(i) \,\big\} \big) \Big],
\end{aligned}
\label{eq:node_update}
\end{equation}

\begin{equation}
\mathbf{e}_{ji}^{(l)} = \mathbf{W}_{\alpha}^{(l-1)}\cdot \mathbf{P}^{(l)}  [\mathbf{h}_j^{(l-1)} \| \mathbf{h}_i^{(l-1)} \| \mathbf{e}_{ji}^{(l-1)}]
,
\label{eq:edge_update}
\end{equation}
where $\mathbf{U}, \mathbf{W}, \mathbf{O}$, and $\mathbf{P}$ are learnable parameters, $[\cdot\| \cdot]$ is the concatenation, and $\mathbf{W_{\alpha}}$ represents the edge weight.

\subsection{Causal-Biased Disentanglement Module}


The proposed CBDM explicitly disentangles the causal and biased features in Eq.~\ref{disentanglement}. This design serves as a simple extension to the baseline model, by replicating the message-passing network and inserting a learnable gating mechanism to assign each edge a soft weight for causal or biased branches. The two GNN branches operate in parallel and share the same initial input graph, node vectors, and edge embeddings, $\mathcal{G}_c=\mathcal{G}_b=\mathcal{G} = (\mathcal{V}, \mathcal{E})$. Random edge dropout is applied to both graphs.

\subsubsection{Dual-branch Graph Neural Network}
We adopt the baseline GNN to construct two GNNs for causal features and biased features separately. To enable the dual GNNs to extract disentangled features from the network, we introduce a gating function that softly assigns each edge to the causal or biased branch. For each edge $(j \rightarrow i)$ with node features $\mathbf{h}_j$, $\mathbf{h}_i$ and edge feature $\mathbf{e}_{ji}$, we compute:
\begin{equation}
    \scalebox{0.9}{$\displaystyle
\alpha_{ji}^{(l)} = \text{MLP}_{\text{gate}}([\mathbf{h}_j^{(l)} \, \| \, \mathbf{h}_i^{(l)}]), \quad c_{ji}^{(l)} = \sigma(\alpha_{ji}^{(l)}), \quad b_{ji}^{(l)} = 1 - c_{ji}^{(l)},
$}
\end{equation}
where $\text{MLP}_{\text{gate}}$ is the multilayer perceptron, $\sigma$ is the sigmoid activation function, and $c_{ji}^{(l)}, b_{ji}^{(l)} \in [0,1]$ represent the soft weights of the edge contribution in the $l$-th layer. The edge weights are then used to modulate message propagation in each branch following Eq. \ref{eq:edge_update}, where $\mathbf{W_{\alpha}}$ is replaced by $c_{ji}^{(l)}$ or $b_{ji}^{(l)}$, respectively.

Each GNN branch stacks $L$ layers of propagation and produces final representations:
\begin{equation}
    \mathbf{Z}_c = \text{GNN}_c(\mathcal{G}_c), \quad \mathbf{Z}_b = \text{GNN}_b(\mathcal{G}_b).
\label{eq:dualgraph}
\end{equation}

\subsubsection{Uncorrelated Representation Learning}
\label{uncorrelate}
Given the learning representation from the dual-branch GNNs, we need to ensure that the learned causal features and biased features are uncorrelated. Inspired by~\cite{lee2021learning,fan2022debiasing}, we simultaneously train the paired representations $\mathbf{Z}_\text{concat}=[\mathbf{z}_c \| \,\mathbf{z}_b]$ with two linear classifiers. The concatenated representations are passed to two classifiers $f_c(\cdot)$ and $f_b(\cdot)$, where $f_c$ is trained with standard cross-entropy (CE), and $f_b$ with generalized cross-entropy (GCE) \cite{zhang2018generalized} to amplify the bias:
\begin{equation}
\text{GCE}(f_b(\mathbf{Z}_\text{concat}), y) = \frac{1 - f_b^y(\mathbf{Z}_\text{concat})^q}{q},
\end{equation}
where $q \in (0, 1]$ controls the degree of gradient amplification. Unlike CE loss, GCE gives larger gradient to samples with high confidence predictions, which typically come from biased patterns. As a result, the biased GNN branch is driven to focus on learning shortcuts from $\mathbf{Z}_b$. Importantly, the gradients from GCE are not back-propagated to $\text{GNN}_c$ to avoid leakage into $\mathbf{Z}_c$, while the CE loss is not back-propagated to $\text{GNN}_b$. Therefore, the optimization function for disentanglement is:
\begin{equation}
    \mathcal{L}_\text{dis}=\text{CE}(f_c(\mathbf{Z}_\text{concat}),\mathbf{y})+\text{GCE}(f_b(\mathbf{Z}_\text{concat}),\mathbf{y}).
\end{equation}

To further achieve the goal of decorrelating the causal feature and biased feature, we propose to use the combined counterfactual embeddings to decorrelate $\mathbf{Z}_c$ and $\mathbf{Z}_b$. We randomly select the biased feature $\hat{\mathbf{Z}_b}$ from the other patients with label $\hat{\mathbf{y}}$ to combine the causal feature $\mathbf{Z}_c$ as the counterfactual samples $\hat{\mathbf{Z}}_{\text{concat}}=[\mathbf{Z}_c \| \,\hat{\mathbf{Z}_b}]$. With the generated counterfactual samples, we use the following loss function to train the dual-branch GNNs:
\begin{equation}
    \mathcal{L}_\text{counterfactual}=\text{CE}(f_c(\hat{\mathbf{Z}}_\text{concat}),\mathbf{y})+\text{GCE}(f_b(\hat{\mathbf{Z}}_\text{concat}),\hat{\mathbf{y}}).
\end{equation}
The total loss function is:
\begin{equation}
\mathcal{L}_{\text{total}} = \mathcal{L}_{\text{dis}} + \alpha \cdot \mathcal{L}_{\text{counterfactual}},
\label{eq:totalloss}
\end{equation} 
where $\alpha$ is the hyperparameter to weight the importance. During the training process, we only train the model with $\mathcal{L}_{\text{dis}}$ to ensure the model learn well-disentangled embeddings. Afterward, we use the $\mathcal{L}_{\text{total}}$ to train the whole model. The concatenated feature with classifier $f_c(\cdot)$ is used for inference.

\subsection{Missingness Deconfounding Module}

The proposed MDM aims to approximate the interventional prediction $P(\mathbf{Y} \mid \mathbf{M} ,\textit{do}(\mathbf{X}_O))$ detailed in Eq.~\ref{backdoor}, by explicitly modeling the confounding distribution over modality missingness patterns and marginalizing over it through a learnable attention-based weighting. As illustrated in Fig.~\ref{fig:framework}, MDM is inserted after the dual-branch graph neural network to adjust the generated embeddings.  

\subsubsection{Confounder Dictionary Construction.}
We treat the modality mask pattern as a proxy for the latent confounder. We simulate a set of missingness-induced features by applying random masks to the observed input $\mathbf{X}_O$, generating a masked multimodal dataset $\mathbf{X}'_O$. The new set $\mathbf{X}'_O$ is fed to a pretrained backbone network $\phi(\cdot)$ to obtain the features set $\mathcal{F}_\mathbf{M} = \left\{ \mathbf{m}_k \in \mathbb{R}^d \right\}_{k=1}^{N_k}
$ under different mask patterns $\mathbf{M}$, where $N_k$ is the number of created samples. To compute prototypes, we use the K-Means++ with principal component analysis to learn the \textit{confounder dictionary} $\mathbf{Z}_d=[\mathbf{z}_1, \mathbf{z}_2, ..., \mathbf{z}_K]$, where $K$ is the dictionary size. The prototype feature is computed as the average of each cluster: $\mathbf{z}_i = \frac{1}{N_i^c} \sum_{j=1}^{N_i^c} \mathbf{m}_j^i$, where $N_i^c$ is the number of features in the $i$-th cluster.

\subsubsection{NWGM-Based Interventional Approximation}

To approximate the interventional distribution $P(\mathbf{Y} \mid \textit{do}(\mathbf{X}_O))$, we use the \textit{Normalized Weighted Geometric Mean} (NWGM) \cite{xu2015show} to reduce the complex computation by multiple forward passes of all $\mathbf{z}$. Eq.~\ref{backdoor} can be reformulated as the expectation at the feature level:
\begin{equation}
P(\mathbf{Y} \mid \mathbf{M},\textit{do}(\mathbf{X}_O)) \approx P(\mathbf{Y} \mid  \mathbf{X}_O, \mathbf{C}, \mathbf{B}),
\end{equation}
\begin{equation}
    \mathbf{C}=\sum_\mathbf{z} f_c(\mathbf{X}_O,\mathbf{z})P(\mathbf{z} \mid \mathbf{M}),
\end{equation}
\begin{equation}
    \mathbf{B}=\sum_\mathbf{z} f_b(\mathbf{X}_O,\mathbf{z})P(\mathbf{z} \mid \mathbf{M}).
\end{equation}

Inspired by \cite{wang2020visual}, we use a network to approximate the conditional probability as:

\begin{equation}
P(\mathbf{Y} \mid \mathbf{M},\textit{do}(\mathbf{X}_O)) = \mathbf{W}_h \mathbf{Z} + \mathbf{W}_g \mathbb{E}_{\mathbf{z}\sim P(\mathbf{z}|\mathbf{M})}[g(\mathbf{z})],
\label{eq:NWGM}
\end{equation}
where $W_h \in \mathbb{R}^{d_m \times d}$ and $W_g \in \mathbb{R}^{d_m \times d}$ are learnable parameters, and $\mathbf{Z}$ represents the causal feature $\mathbf{Z}_c$ or biased feature $\mathbf{Z}_b$. According to the prototypes that we have achieved, we approximate $\mathbb{E}_{\mathbf{z}\sim P(\mathbf{z}|\mathbf{M})}[g(\mathbf{z})]$ as:
\begin{equation}
    \mathbb{E}_{\mathbf{z}\sim P(\mathbf{z}|\mathbf{M})}[g(\mathbf{z})]\approx \sum_{i=1}^{N} \mathbf{\lambda}_i \mathbf{z}_i P(\mathbf{z}_i),
\label{eq:gz}
\end{equation}
where $P(\mathbf{z}_i)=\frac{N^c_i}{N_k}$, and $\mathbf{\lambda}_i$ is the coefficient computed by dot product attention:
\begin{equation}
\mathbf{\lambda}_i = \text{softmax}\left( \frac{(\mathbf{W}_q \mathbf{h})^\top (\mathbf{W}_k \mathbf{z}_i)}{\sqrt{d}} \right),
\end{equation}
where $\mathbf{W}_q \in \mathbb{R}^{d_n \times d}$, and $\mathbf{W}_k \in \mathbb{R}^{d_n \times d}$ are learnable layers.

\begin{algorithm}[t]
\caption{Training and Inference of the CaD.}
\label{algorithm}
\KwIn{Training data $\mathcal{D}_{tr}$, label $\mathbf{y}$}
\KwOut{Trained prediction model}

\textbf{Training:} \\
\For{each epoch}{
    \For{each batch in $\mathcal{D}_{tr}$}{
        Sample mini-batch $(\mathbf{X}_O, \mathbf{M}, \mathbf{y})$ \\
        Encode edge embeddings with $\text{Encoder}_m(\mathbf{x}_m^{(p)})$ by Eq. \ref{eq:encoder} \\
        Build bipartite patient-modality graph $\mathcal{G}$ by Eq. \ref{eq:graph}\\
        \textit{// Causal-Biased Disentanglement} \\
        Use edge dropout with to generate bipartite patient-modality graphs $\mathcal{G}_c$ and $\mathcal{G}_b$ \\
        Apply dual-branch GNN with edge-wise gating to obtain $\mathbf{Z}_c$, $\mathbf{Z}_b$ by Eq. \ref{eq:dualgraph}\\

        \textit{// Missingness Deconfounding} \\
        Use pretrained $\phi(\cdot)$ on masked samples to build confounder dictionary $\mathbf{Z}_d$ \\
        Compute weighted vector $\mathbb{E}[g(\mathbf{z})]$ via NWGM by Eq. \ref{eq:gz}\\
        Generate deconfounded embeddings $\mathbf{Z}'_c$, $\mathbf{Z}'_b$ with NWGM by Eq. \ref{eq:NWGM}\\
        Construct concatenated representation ${\mathbf{Z}}'_{\text{concat}} = [\mathbf{Z}'_c \| \mathbf{Z}'_b]$  \\
        Construct counterfactual representation $\hat{\mathbf{Z}}'_{\text{concat}} = [\mathbf{Z}'_c \| \hat{\mathbf{Z}}'_b]$ \\
        Compute prediction loss $\mathcal{L}_{\text{total}}$ by Eq. \ref{eq:totalloss} and update parameters \\
    \textbf{end for}
    }
\textbf{end for}}

\vspace{0.5em}
\textbf{Inference:} \\
\For{each test batch $(\mathbf{X}_O, \mathbf{M})$}{
    Build graph $\mathcal{G}$ and extract embeddings $\mathbf{Z}'_c$, $\mathbf{Z}'_b$ \\
    Compute NWGM-based correction and obtain ${\mathbf{Z}}'_{\text{concat}}$ \\
    Predict label via $f_c({\mathbf{Z}}'_{\text{concat}})$ \\
    \textbf{end for}
}
\end{algorithm}

\subsection{Final Implementation}
Both the causal feature $\mathbf{Z}_c$ and biased feature $\mathbf{Z}_b$ are input into MDM to generate missingness unbiased features $\mathbf{Z}'_c$ and $\mathbf{Z}'_b$. In order to decorrelate the causal feature and biased feature, the obtained $\mathbf{Z}'_c$ and $\mathbf{Z}'_b$ are concatenated as ${\mathbf{Z}}'_\text{concat}$, while $\mathbf{Z}'_c$ and $\hat{\mathbf{Z}}'_b$ are concatenated as counterfactual samples $\hat{{\mathbf{Z}}}'_\text{concat}$ to replace $\mathbf{Z}_{\text{concat}}$ and $\hat{\mathbf{Z}}_{\text{concat}}$ in Sec.~\ref{uncorrelate} for training. The summary of the proposed \textbf{CaD} is presented in Algorithm \ref{algorithm}.

\section{Experiments}~\label{sec5}

\subsection{Datasets and Task Description}
\label{datasetandtask}

We evaluate our method on four real-world clinical datasets: MIMIC-IV \cite{johnson2023mimic}, eICU \cite{pollard2018eicu}, ADNI \cite{jack2008alzheimer}, and an in-hospital atrial fibrillation (AFib) dataset. Each dataset presents unique challenges in terms of the lack of modality and the heterogeneity of the results. We follow the same setting as in \cite{wu2024multimodal} for experiments on the three public datasets. The detailed description of the datasets is included in Sec. I of  \textit{Supplementary Material}.

\subsection{Baseline Models}
We compare our method with seven representative baselines that span three categories: (1) \textbf{modality imputation methods} such as CM-AE \cite{ngiam2011multimodal} and SMIL \cite{ma2021smil}; (2) \textbf{direct prediction methods} including MT \cite{ma2022multimodal}, GRAPE \cite{you2020handling}, HGMF \cite{chen2020hgmf}, M3Care \cite{zhang2022m3care}, and MUSE \cite{wu2024multimodal}. In addition, we integrate our causal debiasing framework with MT, GRAPE, M3Care, and MUSE to evaluate its compatibility and effectiveness. Full descriptions of all baselines are provided in Sec. II of  \textit{Supplementary Material}.

\subsection{Implementation Details}
\label{implementation}
\textbf{Confounder Setup.} To construct the latent confounder dictionary, we randomly mask out one modality to generate a missingness-conditioned feature vector \(\mathbf{m}\), which serves as a proxy for confounder semantics. For each training sample with no missing modalities, we use the pre-trained GRAPE \cite{you2020handling} as the backbone and directly extract \(\mathbf{m}\) from its embedding space. Thus, the dimensionality of \(\mathbf{m}_k\) and the confounder embedding \(\mathbf{z} \in \mathbb{R}^d\) is aligned with the baseline model's embedding dimension \(d = 128\). The number of clusters \(K\) is set to \(K = 64\) for MIMIC-IV and eICU, \(K = 32\) for ADNI, and \(K = 16\) for AFib.

\textbf{Training Details.} All experiments are conducted using the PyTorch framework on one NVIDIA A100 GPU. To ensure a fair comparison, we closely follow the experimental protocol of the MUSE baseline. Specifically, for the MIMIC-IV, eICU, and ADNI datasets, we adopt the same loss functions, learning rate, and batch size settings as reported in the original MUSE implementation. For the AFib dataset, additional training configurations are provided in Sec. II of  \textit{Supplementary Material}. We set the degree of gradient amplification $q$ at 0.7, the hyperparameter $\alpha$ to be 0.5, and the dimensions of the hidden projection layers at \(d_m = 128\) and \(d_n = 64\) in all experiments. We train the model with $\mathcal{L}_\text{dis}$ for 15 epochs and then train the whole model with $\mathcal{L}_\text{total}$ until 100 epochs. The random edge dropout rate is 10\%. For each metric, we report the mean and standard deviation computed over 1000 bootstrap resamples with replacement. Additionally, we perform independent two-sample t-tests to evaluate whether the improvements achieved by CaD over the baselines are statistically significant.

\subsection{Main Results}
\label{mainresults}

\subsubsection{Results on the ICU Datasets} As shown in Table~\ref{tab:icu}, we observe that MUSE+CaD achieves the new SOTA. Note that we do not include the comparison of MUSE+ in \cite{wu2024multimodal}, as it explores missing labels with semi-supervised learning, which is different from our research topic. Specifically, the CaD-based MT, GRAPE, M3Care, and MUSE improve the AUC-PRC scores on MIMIC-IV by 2.85\%, 3.09\%, 2.95\%, and 2.01\% respectively. CaD also brings noticeable improvements to the eICU dataset with improvements of 3.19\%, 3.04\% and 2.79\% with GRAPE, M3Care and MUSE in AUC-PRC for mortality prediction. 

\begin{table*}[t]
    \centering
    \caption{Results on ICU datasets. A dagger ($\dagger$) indicates the standard deviation is greater than 0.02. An asterisk (*) indicates that the CaD-based models achieves a significant improvement over the baseline methods, with a p-value smaller than 0.05.}
    \renewcommand{\arraystretch}{1}
    \setlength{\tabcolsep}{9.5pt}
    \begin{tabular}{lcccccccc}
        \toprule
        \multirow{3.7}{*}{\centering \textbf{Method}} & \multicolumn{4}{c}{\textbf{MIMIC-IV}} & \multicolumn{4}{c}{\textbf{eICU}} \\
        \cmidrule(lr){2-5} \cmidrule(lr){6-9}
        & \multicolumn{2}{c}{\textbf{Mortality}} & \multicolumn{2}{c}{\textbf{Readmission}} & \multicolumn{2}{c}{\textbf{Mortality}} & \multicolumn{2}{c}{\textbf{Readmission}} \\
        \cmidrule(lr){2-3} \cmidrule(lr){4-5} \cmidrule(lr){6-7} \cmidrule(lr){8-9}
        & \textbf{AUC-ROC} & \textbf{AUC-PRC} & \textbf{AUC-ROC} & \textbf{AUC-PRC} & \textbf{AUC-ROC} & \textbf{AUC-PRC} & \textbf{AUC-ROC} & \textbf{AUC-PRC} \\ 
        \midrule
        CM-AE \cite{ngiam2011multimodal}   & 0.8530$^\dagger$ & 0.4351$^\dagger$ & 0.6817$^\dagger$ & 0.4324$^\dagger$ & 0.8624 & 0.3902 & 0.7462$^\dagger$ & 0.4338$^\dagger$ \\
        SMIL \cite{ma2021smil}   & 0.8607 & 0.4438 & 0.6894$^\dagger$ & 0.4368$^\dagger$ & 0.8711 & 0.4066 & 0.7506 & 0.4447 \\
        MT \cite{ma2022multimodal}    & 0.8739 & 0.4452 & 0.6901 & 0.4375 & 0.8882 & 0.4109 & 0.7635 & 0.4500 \\
        GRAPE \cite{you2020handling}   & 0.8837 & 0.4584$^\dagger$ & 0.7085 & 0.4551 & 0.8903 & 0.4137 & 0.7663 & 0.4501 \\
        HGMF \cite{chen2020hgmf}   & 0.8710 & 0.4433 & 0.7005$^\dagger$ & 0.4421 & 0.8878 & 0.4104 & 0.7604 & 0.4496$^\dagger$ \\
        M3Care \cite{zhang2022m3care}  & 0.8896$^\dagger$ & 0.4603$^\dagger$ & 0.7067 & 0.4532 & 0.8964 & 0.4155 & 0.7598$^\dagger$ & 0.4430 \\
        MUSE \cite{wu2024multimodal}  & 0.9004 & 0.4735 & 0.7152 & 0.4670$^\dagger$ & 0.9017 & 0.4216 & 0.7709 & 0.4631 \\
        \midrule
        \textbf{MT+CaD}   & 0.8973* & 0.4737* & 0.7108* & 0.4497* & 0.9017* & 0.4186* & 0.7715* & 0.4587* \\
        \textbf{GRAPE+CaD}   & 0.9235* & 0.4893* & 0.7425* & 0.5023* & 0.9421* & 0.4456* & 0.7963* & 0.4802* \\
        \textbf{M3Care+CaD}   & 0.9247* & 0.4898* & 0.7409* & 0.5010* & 0.9427* & 0.4459* & 0.7896*$^\dagger$ & 0.4742* \\
        \textbf{MUSE+CaD}   & \textbf{0.9283*} & \textbf{0.4936*} & \textbf{0.7489*} & \textbf{0.5078*} & \textbf{0.9476*} & \textbf{0.4495*} & \textbf{0.8022*} & \textbf{0.4887*} \\
        \bottomrule
    \end{tabular}
    \label{tab:icu}
\end{table*}

\begin{table}[t]
\centering
\caption{Performance comparison on ADNI and AFib datasets. ``--'' indicates that the method was not evaluated on that task. The dagger ($\dagger$) and asterisk (*) have the same meanings as Tab.~\ref{tab:icu}.}
\vspace{5pt}
\label{tab:adni_afib}
\renewcommand{\arraystretch}{1.1}
\setlength{\tabcolsep}{3pt}
\resizebox{\linewidth}{!}{
\begin{tabular}{lcccccc}
\toprule
\multirow{2}{*}{\textbf{Method}} 
& \multicolumn{2}{c}{\textbf{ADNI (AD Progression)}} 
& \multicolumn{2}{c}{\textbf{AFib (Classification)}} 
& \multicolumn{2}{c}{\textbf{AFib (Recurrence)}} \\
\cmidrule(lr){2-3} \cmidrule(lr){4-5} \cmidrule(lr){6-7}
 & \textbf{AUC-ROC} & \textbf{Accuracy} & \textbf{AUC-ROC} & \textbf{Accuracy} & \textbf{AUC-ROC} & \textbf{Accuracy} \\
\midrule
MLP           & --     & --     & 0.6863$^\dagger$ & 0.5304$^\dagger$ & 0.6342$^\dagger$ & 0.5108$^\dagger$ \\
CM-AE \cite{ngiam2011multimodal}       & 0.8722$^\dagger$ & 0.7305$^\dagger$ & --     & --     & --     & -- \\
SMIL \cite{ma2021smil}        & 0.8761$^\dagger$ & 0.7338$^\dagger$ & 0.7895$^\dagger$ & 0.6654$^\dagger$ & 0.7356$^\dagger$ & 0.6178$^\dagger$ \\
MT \cite{ma2022multimodal}
& 0.8935           & 0.7604           & 0.8034 & 0.6849 & 0.7623$^\dagger$ & 0.6470$^\dagger$ \\
GRAPE \cite{you2020handling}      & 0.9031$^\dagger$ & 0.7820$^\dagger$ & 0.8246$^\dagger$ & 0.6998$^\dagger$ & 0.7958$^\dagger$ & 0.6723$^\dagger$ \\
HGMF \cite{chen2020hgmf}        & 0.8845$^\dagger$ & 0.7463$^\dagger$ & --     & --     & --     & -- \\
M3Care \cite{zhang2022m3care}      & 0.9101           & 0.7822           & --     & --     & --     & -- \\
MUSE \cite{wu2024multimodal}       & 0.9158$^{\dagger}$ & 0.7973$^{\dagger}$ & 0.8327$^\dagger$ & 0.7123$^\dagger$ & 0.8075$^\dagger$ & 0.6833$^\dagger$ \\
\midrule
\textbf{MT+CaD}       & 0.9106$^{*}$     & 0.7818$^{*}$     & 0.8176$^{*}$ & 0.6917$^{*}$ & 0.7767$^{*}$ & 0.6533$^{*}$ \\
\textbf{GRAPE+CaD}    & 0.9345$^{*}$    & 0.8358$^{*}$     & 0.8562${*\dagger}$ & 0.7255${*\dagger}$ & 0.8129${*\dagger}$ & 0.6855${*\dagger}$ \\
\textbf{M3Care+CaD}   & 0.9356$^{*}$     & 0.8339$^{*}$     & --     & --     & --     & -- \\
\textbf{MUSE+CaD}     & \textbf{0.9415}$^{*}$ & \textbf{0.8388}$^{*}$ & \textbf{0.8593}$^{*}$ & \textbf{0.7329}$^{*}$ & \textbf{0.8175}$^{*}$ & \textbf{0.6898}$^{*}$ \\
\bottomrule
\end{tabular}
}
\end{table}

\subsubsection{Results on the ADNI Dataset}
Table~\ref{tab:adni_afib} presents results on Alzheimer's disease prediction using the ADNI dataset. Compared to their baselines, GRAPE+CaD and M3Care+CaD improve AUC-ROC by 3.14\% and 2.55\%, while MUSE+CaD achieves an absolute gain of 2.57\% in AUC-ROC and 4.15\% in accuracy. These results confirm the benefit of CaD even in smaller datasets, due to its ability to mitigate the effect of population shift and modality-specific missingness.

\subsubsection{Results on the AFib Dataset}
As shown in Table~\ref{tab:adni_afib}, the improvements on the AFib dataset are relatively modest. Moreover, M3Care \cite{zhang2022m3care} and HGMF \cite{chen2020hgmf} cannot converge during training. MLP denotes using multilayer perceptron to process the contatenated multimodal features. For classification, CaD improves AUC-ROC by 3.16\% (GRAPE) and 2.66\% (MUSE). For recurrence prediction, improvements are 1.71\% and 1.00\%. The performance gain on the AFib dataset is comparatively smaller with larger deviation. This can be explained by the limited data size, which results in fewer observable confounding effects. However, CaD-based models still outperform their baselines.

\subsection{Ablation Study}

\begin{table*}[t]
\centering
\caption{Ablation study on the influence of each module and variant.}
\label{tab:ablation_main}
\setlength{\tabcolsep}{8.6pt}
\renewcommand{\arraystretch}{1}
\begin{tabular}{llccccc}
\toprule
\multirow{2}{*}{\textbf{ID}} & \multirow{2}{*}{\textbf{Method}} & \multicolumn{2}{c}{\textbf{MIMIC-IV}} & \multicolumn{2}{c}{\textbf{eICU}} & \multicolumn{1}{c}{\textbf{ADNI}} \\
\cmidrule(lr){3-4} \cmidrule(lr){5-6} \cmidrule(lr){7-7}
 &  & \textbf{Mortality} & \textbf{Readmission} & \textbf{Mortality} & \textbf{Readmission} & \textbf{AD Progression} \\
\midrule
A1.1 & GRAPE+CaD w/o CBDM & 0.8943 & 0.7137 & 0.9033 & 0.7725 & 0.9129 \\
A1.2 & MUSE+CaD w/o CBDM & 0.9087 & 0.7266 & 0.9153 & 0.7778 & 0.9267 \\
A2.1 & GRAPE+CaD w/o counterfactual supervision & 0.9129 & 0.7288 & 0.9318 & 0.7842 & 0.9209 \\
A2.2 & MUSE+CaD w/o counterfactual supervision & 0.9207 & 0.7407 & 0.9324 & 0.7938 & 0.9286 \\
A3.1 & GRAPE+CaD w/ InfoNCE + Entropy Max. & 0.9112 & 0.7253 & 0.9278 & 0.7799 & 0.9223 \\
A3.2 & MUSE+CaD w/ InfoNCE + Entropy Max. & 0.9218 & 0.7423 & 0.9328 & 0.7966 & 0.9275 \\
\midrule
A4.1 & GRAPE + CaD w/o MDM & 0.9159 & 0.7328 & 0.9314 & 0.7846 & 0.9238 \\
A4.2 & MUSE+CaD w/o MDM & 0.9230 & 0.7435 & 0.9397 & 0.7928 & 0.9369 \\
A5.1 & GRAPE+CaD w/ multi-round masking & 0.9287 & 0.7466 & 0.9456 & 0.7969 & 0.9352 \\
A5.2 & MUSE+CaD w/ multi-round masking & 0.9292 & 0.7493 & 0.9522 & 0.8058 & 0.9419 \\
A6.1 & GRAPE+CaD w/ random $Z$ & 0.9108 & 0.7280 & 0.9297 & 0.7832 & 0.9210 \\
A6.2 & MUSE+CaD w/ random $Z$ & 0.9226 & 0.7442 & 0.9357 & 0.7913 & 0.9307 \\
A7.1 & GRAPE+CaD w/ MT & 0.9185 & 0.7376 & 0.9364 & 0.7923 & 0.9258 \\
A7.2 & MUSE+CaD w/ MT & 0.9267 & 0.7470 & 0.9441 & 0.8003 & 0.9393 \\
A8.1 & GRAPE+CaD w/ MUSE & 0.9257 & 0.7431 & 0.9400 & 0.7949 & 0.9356 \\
A8.2 & MUSE+CaD w/ MUSE & 0.9310 & 0.7523 & 0.9506 & 0.8038 & 0.9425 \\
\midrule
-- & GRAPE+CaD & 0.9235 & 0.7425 & 0.9421 & 0.7963 & 0.9345 \\
-- & MUSE+CaD & 0.9283 & 0.7489 & 0.9476 & 0.8022 & 0.9415 \\
\bottomrule
\end{tabular}
\label{tab:ablation}
\end{table*}

We conduct ablation studies in Table \ref{tab:ablation} to evaluate the effectiveness of the proposed CaD framework. The experiments are performed on the MIMIC-IV, eICU, and ADNI datasets, using GRAPE \cite{you2020handling} and MUSE \cite{wu2024multimodal} as baseline models.

\subsubsection{Effectiveness of Each Module}
We evaluate the contribution of each CaD module by comparing ablated variants. Removing CBDM (A1.1 and A1.2) leads to a performance drop of 2.9\% (GRAPE) and 2.4\% (MUSE) on MIMIC-IV Readmission, indicating that CBDM plays a critical role by mitigating distributional bias from biased features. In comparison, removing MDM (A4.1 and A4.2) results in smaller drops of 1.0\% and 0.5\%, as the overall missingness rate in the datasets is relatively low, limiting the marginal benefit of modeling missing modality confounding. We make further analysis of effect of MDM with different missing rates.

\subsubsection{Effect of Loss Functions}
In experiments A2.1 and A2.2, we compare the proposed method with and without counterfactual supervision. The results show that the inclusion of counterfactual training leads to measurable improvements, particularly on larger datasets where the data contain more irrelevant features. For example, we observe an obvious performance drop without counterfactual guidance on MIMIC-IV readmission task.

In experiments A3.1 and A3.2, we explore alternative loss designs for disentanglement. Specifically, we use InfoNCE \cite{oord2018representation} to reduce the mutual information between $\mathbf{Z}_c$ and $\mathbf{Z}_b$, and entropy maximization \cite{kim2019learning} to discourage $\mathbf{Z}_b$ from making accurate predictions. This combination can be regarded as similar to the setting in A2.1 and A2.2. Although these losses achieve comparable performance in some cases, we adopt the proposed loss function throughout the paper, given the simplicity and effectiveness of using counterfactual training.

\subsubsection{Effect of Different Masking Strategies}
We evaluate a multi-round masking variant (A5.1 and A5.2), which masks each modality individually across $N$ rounds when $N$ modalities are available, and then computes the final confounder representation by averaging $\mathbf{Z}_i$ over all masked samples. As shown in Table \ref{tab:ablation}, even though multi-round masking achieves marginal improvements, However, given the additional computational overhead from multiple forward passes, we adopt the single-round random masking strategy in our final implementation. Furthermore, we also compare performance using a random dictionary (A6.1 and A6.2) which is not the average of $\mathbf{m}_i$. The experimental results show that the random dictionary would significantly affect performance.

\subsubsection{Robustness of Pre-trained Confounder Backbone}
We evaluate different choices for constructing the confounder dictionary by comparing using pre-trained GRAPE with variants using MT (A7.1 and A7.2) and MUSE (A8.1 and A8.2) as pre-trained encoders. The marginal changes of performance indicate that MDM is not dependent on a well-chosen pre-trained backbone. Moreover, while MUSE offer slight improvements, we adopt GRAPE as the default backbone for confounder construction in our final CaD design, considering the larger model size and training cost.

\subsubsection{Effect of Confounder Size}
\begin{figure*}[t]
    \centering
    \includegraphics[width=\linewidth]{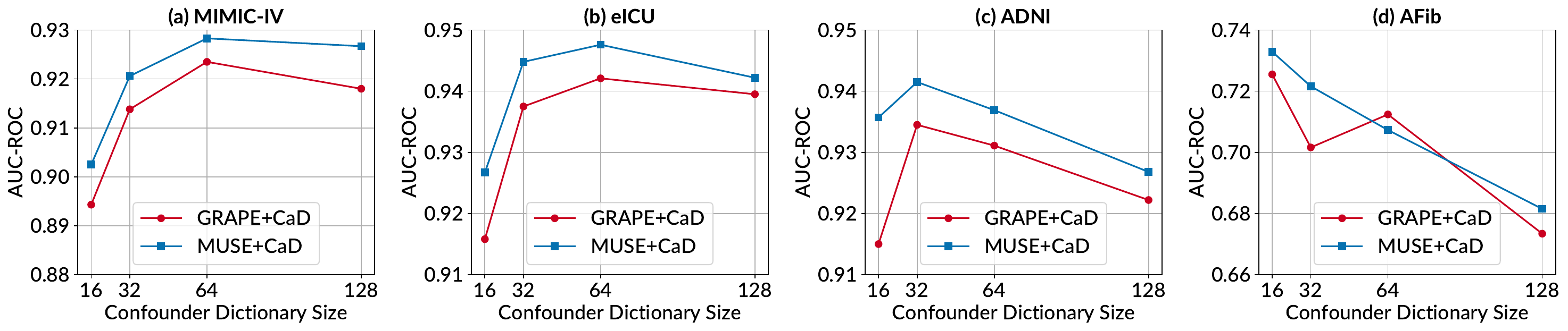}
    \caption{Ablation study for the confounder dictionary size $K$ on four datasets. 
    (a), (b), (c), and (d) represent the results from MIMIC-IV, eICU, ADNI, and AFib.}
    \label{fig:confounder_size}
\end{figure*}

To assess the effect of the size of the confounder dictionary $K$, we set $K$ on 16, 32, 64, 128 and evaluate the performance on MIMIC-IV mortality, eICU mortality, ADNI and the AFib classification task with AUC-ROC reported. As shown in Fig. \ref{fig:confounder_size}, a proper choice of $K$ can significantly affect performance, particularly on larger datasets such as MIMIC-IV and eICU where confounding signals are stronger. Both GRAPE+CaD and MUSE+CaD exhibit the best performance at $K=64$, while the performance decreases to a larger size in smaller datasets ADNI and AFib, probably due to limited patterns inside. 

\subsubsection{Effect of Hyperparameter $\alpha$}

\begin{table}[t]
\centering
\caption{Ablation study on the influence of parameter $\alpha$.}
\label{tab:parameter}
\resizebox{.5\textwidth}{!}{
\begin{tabular}{llcccccc}
\toprule
\multirow{2}{*}{\textbf{Method}} & \multirow{2}{*}{$\alpha$} & \multicolumn{2}{c}{\textbf{MIMIC-IV}} & \multicolumn{2}{c}{\textbf{eICU}} & \multicolumn{1}{c}{\textbf{ADNI}} \\
\cmidrule(lr){3-4} \cmidrule(lr){5-6} \cmidrule(lr){7-7}
 &  & \textbf{Mortality} & \textbf{Readmission} & \textbf{Mortality} & \textbf{Readmission} & \textbf{AD Progression} \\
\midrule
\multirow{4}{*}{GRAPE+CaD} 
 & 0.1 & 0.9156 & 0.7320 & 0.9364 & 0.7893 & 0.9266 \\
 & 0.3 & 0.9202 & 0.7393 & 0.9387 & 0.7928 & 0.9311 \\
 & 0.5 & \textbf{0.9235} & \textbf{0.7425} & \textbf{0.9421} & \textbf{0.7963} & \textbf{0.9345} \\
 & 0.7 & 0.9317 & 0.7401 & 0.9405 & 0.7947 & 0.9298 \\
\midrule
\multirow{4}{*}{MUSE+CaD} 
 & 0.1 & 0.9248 & 0.7432 & 0.9346 & 0.7967 & 0.9354 \\
 & 0.3 & 0.9258 & 0.7470 & 0.9439 & 0.7994 & 0.9393 \\
 & 0.5 & \textbf{0.9283} & 0.7489 & \textbf{0.9476} & \textbf{0.8022} & \textbf{0.9415} \\
 & 0.7 & 0.9276 & \textbf{0.7501} & 0.9420 & 0.7985 & 0.9411 \\
\bottomrule
\end{tabular}
}
\label{tab:appendixparameter}
\end{table}
We conduct an ablation study to investigate the effect of the hyperparameter $\alpha$, which balances the loss contributions in our CaD framework. As shown in Table~\ref{tab:appendixparameter}, we evaluate GRAPE+CaD and MUSE+CaD across three datasets: MIMIC-IV, eICU, and ADNI. We observe that while performance varies slightly with different $\alpha$ values, setting $\alpha = 0.5$ consistently yields the best or near-best results across all tasks. Although minor fluctuations exist (e.g., $\alpha = 0.3$ achieves marginally better results on certain metrics), we choose $\alpha = 0.5$ for all experiments to ensure consistency and stability across settings.

\subsubsection{Embedding Analysis}

\begin{table}[t]
\centering
\caption{Euclidean distance of the representations from the same patients with different modalities.}
\label{tab:euclidean_distance}
\setlength{\tabcolsep}{28pt}
\renewcommand{\arraystretch}{1}
\begin{tabular}{lc}
\toprule
\textbf{Model} & \textbf{Euclidean Distance} \\
\midrule
CM-AE         & 0.7352 \\
SMIL          & 0.6337 \\
MT            & 0.5839 \\
GRAPE         & 0.3001 \\
HGMF          & 0.4426 \\
M3Care        & 0.3286 \\
MUSE      & 0.2437 \\
\midrule
MT+CaD        & 0.2790 \\
GRAPE+CaD     & 0.2215 \\
MUSE+CaD      & 0.2178 \\
\bottomrule
\end{tabular}
\end{table}

Following MUSE \cite{wu2024multimodal}, we measure the Euclidean distance between representations of the same patient using full modalities and those with 30\% randomly masked inputs. We report results on the MIMIC-IV test set for mortality prediction. We use the causal features $\mathbf{Z}'_c$ to compute the distance. As shown in Table~\ref{tab:euclidean_distance}, CaD-enhanced models consistently produce distances lower than their base baselines, indicating more robust and modality-agnostic representations. Notably, MUSE+CaD achieves the smallest distance (0.2178), outperforming MUSE (0.2437) and all baselines.

\subsection{Analysis of Different Missing Rates}
\begin{figure*}[t]
    \centering
    \includegraphics[width=0.8\linewidth]{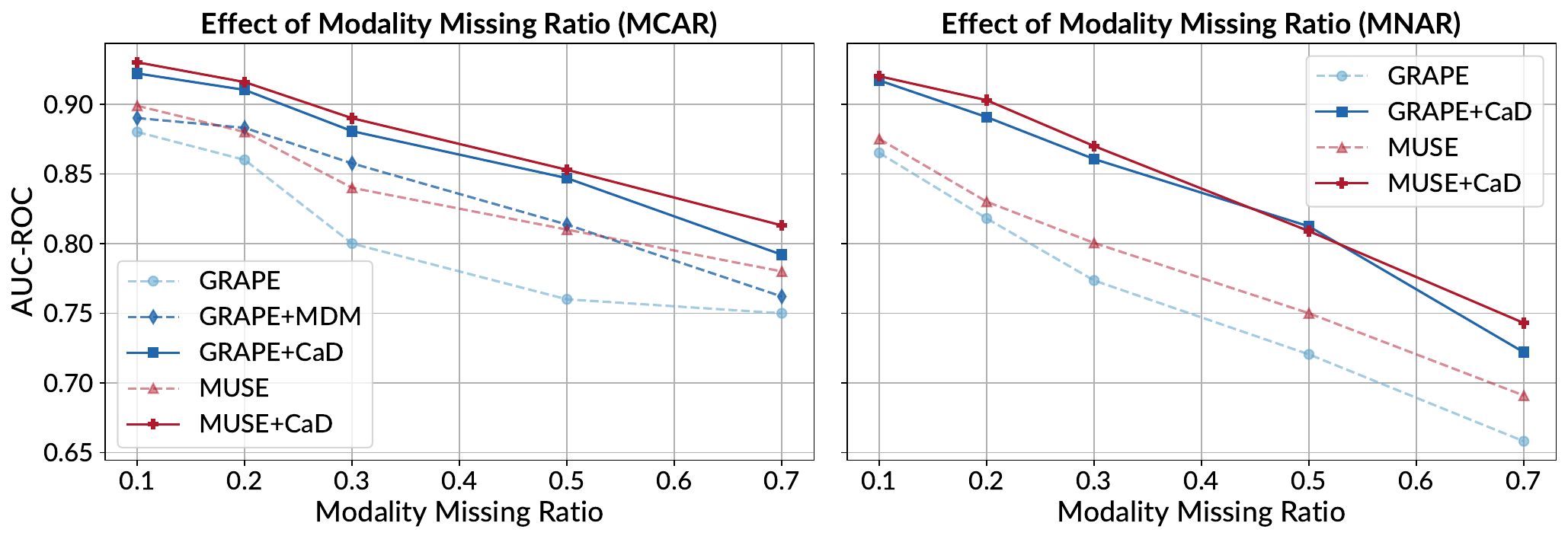}
    \caption{Ablation study for different modality missing rate under different missing patterns (a) MCAR and (b) MNAR.  }
    \label{fig:missingrates}
\end{figure*}

\begin{figure}[t]
    \centering
    \includegraphics[width=0.8\linewidth]{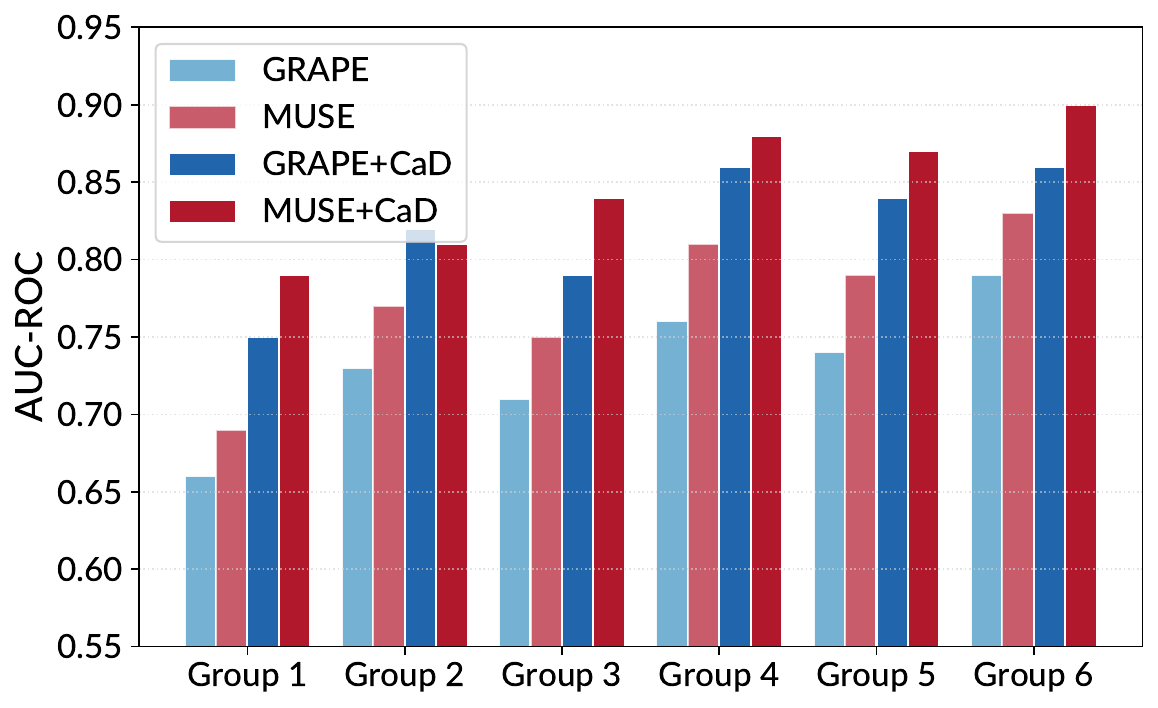}
    \caption{AUC-ROC for different groups under MNAR of different methods in MIMIC-IC dataset.}
    \label{fig:group}
\end{figure}
We perform comparative experiments on the MIMIC-IV mortality prediction task under varying missing ratios of the modality, as illustrated in Fig.~\ref{fig:missingrates}. Under the MCAR setting, we observe that CaD-based models consistently outperform their baselines, particularly when the missing ratio is below 0.5. Additionally, we evaluate GRAPE with the MDM module and find that MDM is more effective as the missing rate increases, supporting its design to handle missingness bias.

Beyond prior work \cite{zhang2022m3care,you2020handling,wu2024multimodal} that mostly assumes the MCAR scenario, we also explore the MNAR setting. Unlike MUSE \cite{wu2024multimodal}, where the MNAR condition is approximated by missing labels, we instead simulate MNAR patterns in a controlled yet clinically plausible way. 
Specifically, we generate the modality mask of four modalities, medication records, lab values, clinical notes, and vital signals, based on two patient attributes: \textbf{age} and \textbf{clinical severity}. We define three age brackets: \textit{young} (age < 40), \textit{middle-aged} (40 $\leq$ age $\leq$ 65), and \textit{old} (age > 65). Clinical severity is determined using the Sequential Organ Failure Assessment (SOFA) score: patients with SOFA < 5 are labeled \textit{low severity}, and those with SOFA $\geq$ 5 are labeled \textit{high severity} \cite{lu2025impact}. Combining these criteria, we construct six patient groups with a template missingness probability $p_{ij}^{\text{template}}$ that reflects structured clinical bias, as shown in Table \ref{template_mnar}, where younger and less severe patients are more likely to have missing data, while older or more severe patients are more likely to be fully observed. To match a target global missing rate, we apply a scaling factor $\alpha$ to the template:
\begin{equation}
    p_{ij}^{\text{final}} = \min(\alpha \cdot p_{ij}^{\text{template}}, 1.0),
\end{equation}
where $p_{ij}^{\text{final}}$ is the final missingness probability for modality $j$ in group $i$. The initial total missing rate after applying the missingness template is $8.7\%$.

The performance of MUSE, GRAPE, MUSE+CaD, and GRAPE+CaD is illustrated in Fig. \ref{fig:missingrates}. The results show that the performance of all methods under MNAR is inferior to the that under MCAR, showing the challenges. However, CaD still provides consistent advantages under MNAR when the missing rate is less than $50\%$, validating its ability to handle both missingness bias and distributional confounding. Finally, we note that GRAPE+CaD and MUSE+CaD achieve comparable performance across missingness levels. We attribute this to the CBDM component in CaD, whose contrastive disentanglement loss function behaves similarly to the mutual-consistent contrastive loss used in MUSE, thus reducing the representation discrepancy between these two methods.

\begin{figure*}[t]
    \centering
    \includegraphics[width=0.85\linewidth]{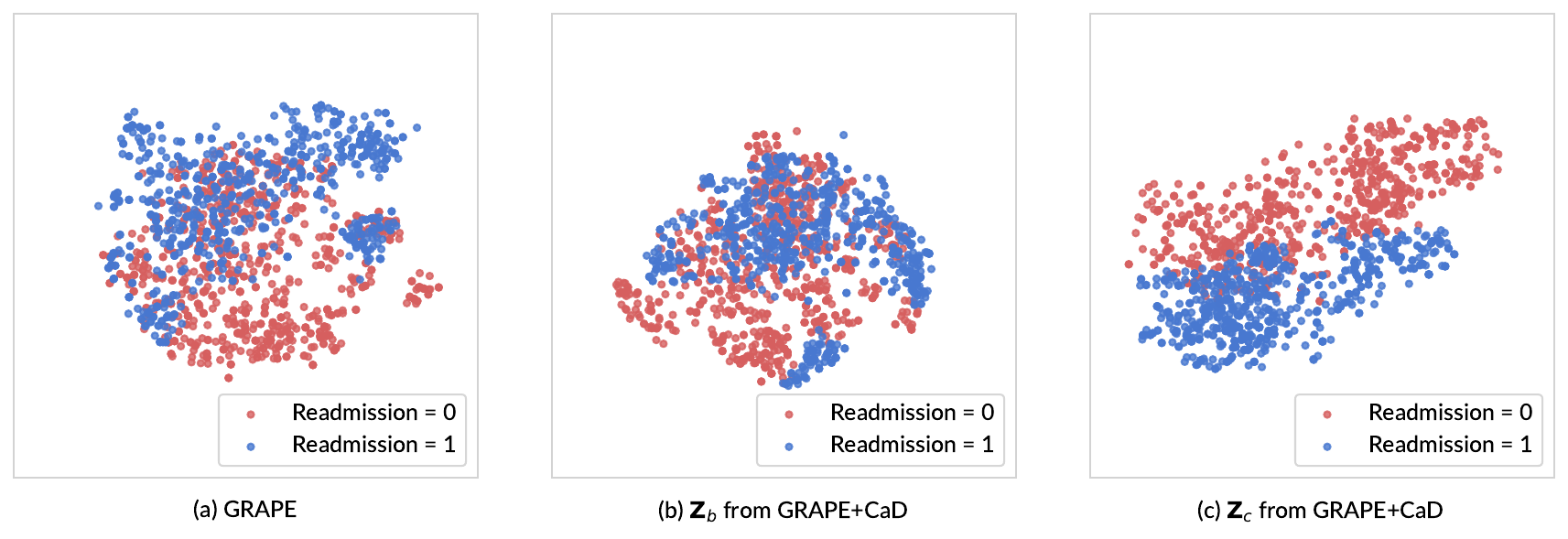}
    \caption{Visualization results of (a) GRAPE, (b) $\textbf{Z}_b$ from GRAPE+CaD, and (c) $\textbf{Z}_c$ from GRAPE+CaD using t-SNE methods on MIMIC-IV Readmission task under MNAR setting with $30\%$ missing ratio.}
    \label{fig:tsne}
\end{figure*}

\begin{table}[t]
\centering
\caption{
Group-conditioned missingness template for MNAR simulation. \textbf{Med}, \textbf{Lab}, \textbf{Notes}, and \textbf{Vital} denote medication records, lab values, clinical notes and vital signals, respectively.}
\label{tab:mnar_template}
\begin{tabular}{clcccc}
\toprule
\textbf{Group} & \textbf{Age, Severity} & \textbf{Med} & \textbf{Lab} & \textbf{Notes} & \textbf{Vital} \\
\midrule
1 & Young, Low severity      & 0.15 & 0.30 & 0.18 & 0.35 \\
2 & Young, High severity     & 0.12 & 0.25 & 0.15 & 0.30 \\
3 & Middle, Low severity     & 0.10 & 0.20 & 0.12 & 0.25 \\
4 & Middle, High severity    & 0.08 & 0.15 & 0.10 & 0.20 \\
5 & Old, Low severity        & 0.05 & 0.10 & 0.08 & 0.15 \\
6 & Old, High severity       & 0.03 & 0.08 & 0.05 & 0.10 \\
\bottomrule
\end{tabular}
\label{template_mnar}
\end{table}

\begin{figure*}[t]
    \centering
    \includegraphics[width=0.85\linewidth]{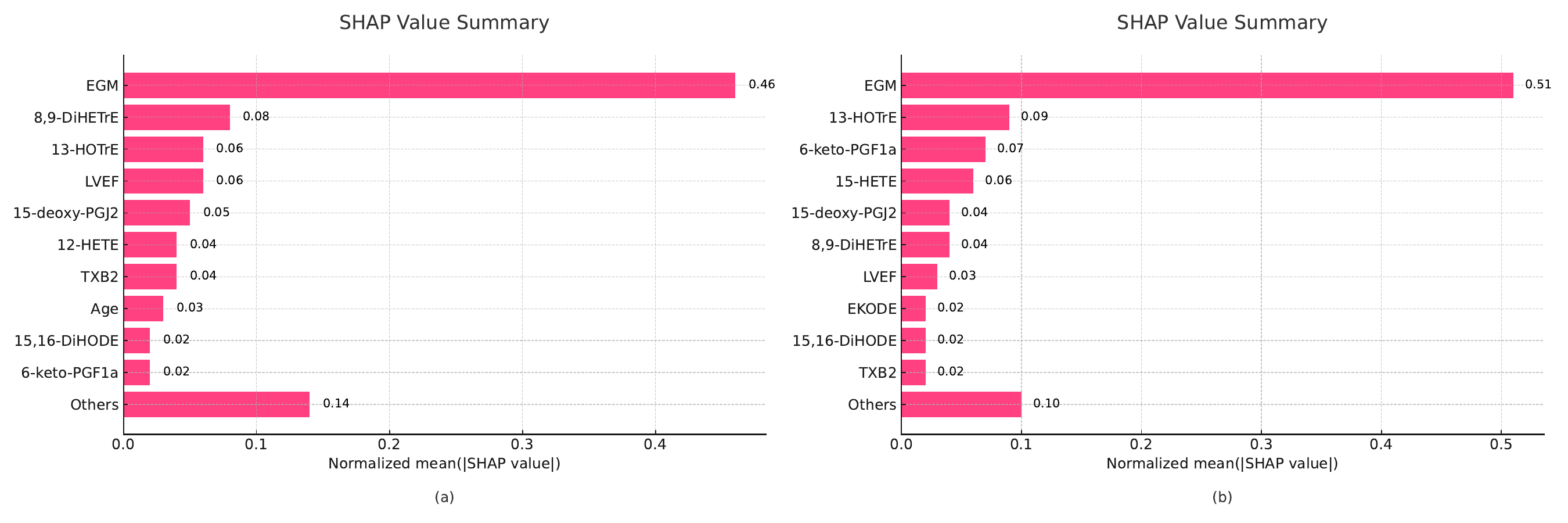}
    \caption{Shapley values comparison of (a) GRAPE and (b) GRAPE+CaD on AFib dataset.}
    \label{fig:shap}
\end{figure*}

\begin{figure}[t]
    \centering
    \includegraphics[width=.9\linewidth]{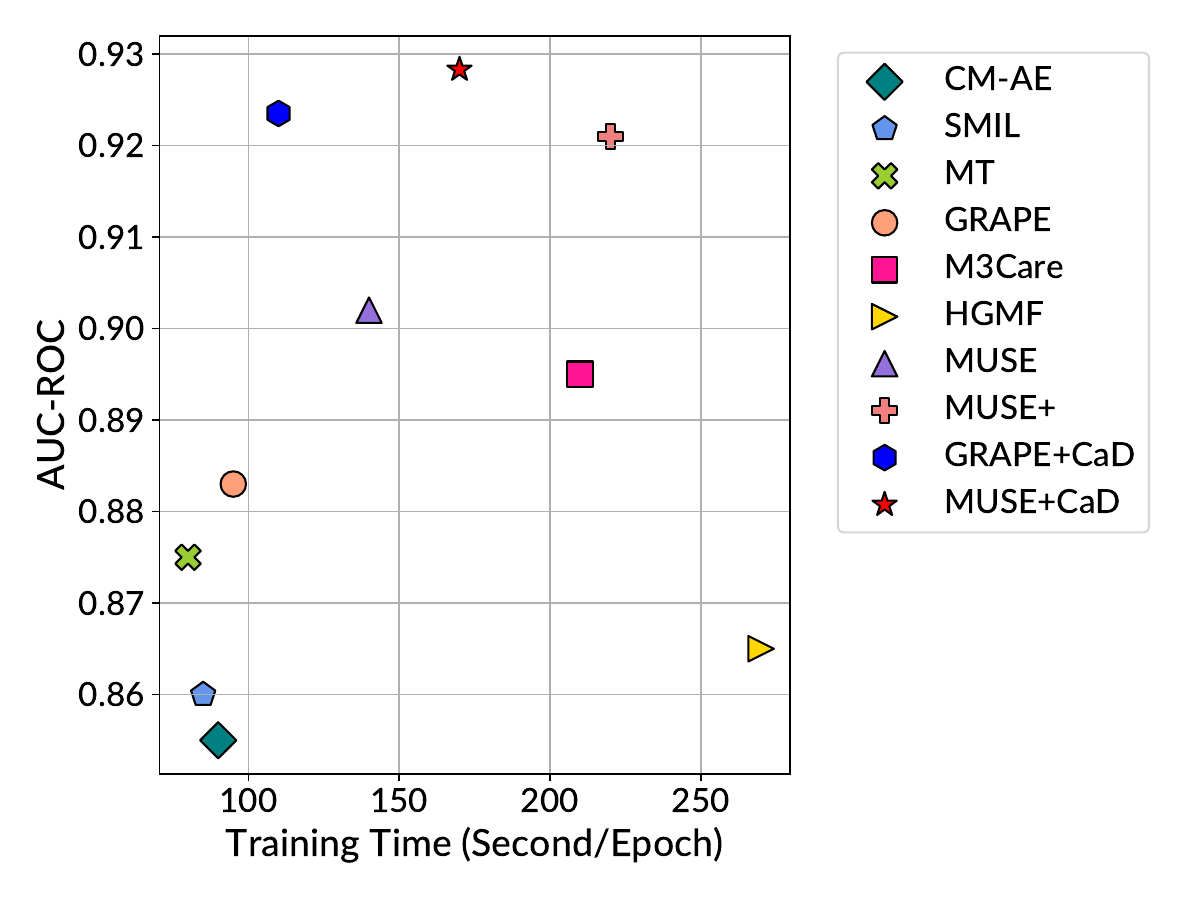}
    \caption{Performance v.s. training time comparisons across different methods.}
    \label{fig:runtime}
\end{figure}

\textbf{Subgroup performance analysis.} Furthermore, we evaluated the robustness of the model in different subgroups of patients with the total missing rate of $30\%$. As shown in Fig. \ref{fig:group}, GRAPE and MUSE exhibit a general performance drop in groups with higher missing modality rates (e.g., Group1 and Group2), reflecting their vulnerability to missing data under MNAR. In contrast, the GRAPE+CaD and MUSE+CaD consistently achieve higher AUC-ROC across all subgroups, particularly outperforming their baselines in high-missingness settings.

\textbf{Visualization results.} Fig.~\ref{fig:tsne} illustrates the t-SNE visualization of patient representations learned by GRAPE and GRAPE+CaD under a 30\% MNAR setting on the MIMIC-IV readmission task. GRAPE shows mixed clusters with little separation between readmission outcomes. In contrast, GRAPE+CaD disentangles \(\mathbf{Z}_b\) (b) and \(\mathbf{Z}_c\) (c), where \(\mathbf{Z}_c\) exhibits clear separation between readmission = 0 and readmission = 1 groups, demonstrating the model's ability to recover the outcome-relevant causal structure.

\textit{Remark 3. Unlike the random edge dropout used in MUSE for contrastive learning, our MDM explicitly models the mechanism of arbitrary modality missingness. By approximating the interventional distribution $p(\mathbf{Y} \mid do(\mathbf{X}_O))$ through weighted aggregation, MDM provides a principled correction for missingness bias, even under MNAR settings.}

\subsection{Shapley Values Analysis}

To understand the contribution of different modalities in the classification of atrial fibrillation, we visualize the normalized Shapley values on the test set for the paroxysmal category. The EGM modality is represented as a 128-dimensional vector, while all other features are scalar clinical or metabolic variables. Fig. \ref{fig:shap} compares the GRAPE baseline (a) and the GRAPE+CaD variant (b). In the baseline model, \texttt{Age} is among the top 10 contributors to the prediction. After applying CaD, \texttt{Age} is excluded from the top 10 features, indicating that our framework effectively reduces the dependence on spurious demographic correlations.

\subsection{Runtime Analysis}
\label{appendix:runtime}

We compare the training time and prediction performance of all methods on the MIMIC-IV mortality task. As shown in Fig. \ref{fig:runtime}, both MUSE+CaD and GRAPE+CaD achieve better AUC-ROC compared to the baseline methods, but with more training time. However, the main computation workload comes from the encoders to extract features from different modalities, whereas the dual-branch graph neural network only increases considerable parameters. In conclusion, the CaD-based methods remain efficient and scalable, demonstrating that our design offers a favorable trade-off between accuracy and training cost.

\section{Conclusion}~\label{sec7}

In this paper, we propose CaD, a causal debiasing framework for medical multimodal representation learning under missing modalities. CaD introduces the Causal-Biased Disentanglement Module and the Missingness Deconfounding Module to tackle the missingness bias and distribution bias, respectively. Extensive experiments on public and in-hospital datasets that CaD consistently improves performance across multiple tasks. This work highlights the importance of incorporating causal principles into multimodal medical representation learning to improve robustness.

\bibliographystyle{IEEEtran}
\bibliography{IEEEabrv,main}

\end{document}


\title{Causal Debiasing Medical Multimodal Representation Learning with Missing Modalities}

\title{Supplementary Material for ``Causal Debiasing Medical Multimodal Representation Learning with Missing Modalities''}

\author{
    Xiaoguang Zhu\orcidlink{0000-0001-9554-2133}, 
    Lianlong Sun\orcidlink{0009-0005-3208-9075}, 
    Yang Liu\orcidlink{0000-0002-1312-0146},
    Pengyi Jiang\orcidlink{0009-0001-7413-9078},
    Uma Srivatsa\orcidlink{0000-0003-0378-2346},
    Nipavan Chiamvimonvat\orcidlink{0000-0001-9499-8817},\\
    Vladimir Filkov\orcidlink{0000-0003-0492-4393}}

\maketitle

\begin{abstract}
We present this supplementary material to accompany our paper, ``Causal Debiasing Medical Multimodal Representation Learning with Missing Modalities''. We created this supplement primarily to adhere to the journal's page limits while still offering a comprehensive resource for readers. In this document, we provide a detailed overview of the datasets in Sec.~\ref{appendix:datasets}. Furthermore, we offer an detailed explaination of the baseline models in Sec.~\ref{appendix:baselines}. Moreover, we also discuss the limitation and future works of the proposed method in Sec.~\ref{appendix:limitation}. All the sections, which are introduced in the main paper, are intended to be a valuable resource for researchers and practitioners seeking a deeper understanding of the paper.
\end{abstract}

\section{Additional Details on the Datasets}
\label{appendix:datasets}
We evaluate our method on three public real-world datasets MIMIC-IV \cite{johnson2023mimic}, eICU \cite{pollard2018eicu} and ADNI \cite{jack2008alzheimer}, following the data preprocessing and data selection protocols of MUSE \cite{wu2024multimodal} to ensure fair comparison. We also verify the proposed method in one in-hospital Atrial Fibrillation (AFib) dataset. All of the datasets are split into 70\%, 10\%, 20\% training, validation, and test sets. Moreover, we made a conditional entropy analysis on the MIMIC-IV and eICU dataset to illustrate how different missing patterns introduce bias to different categories.

\subsection{ICU Datasets}

\textbf{MIMIC-IV.} MIMIC-IV is a publicly available electronic health record (EHR) dataset collected from the Beth Israel Deaconess Medical Center, comprising over 431K ICU admissions from approximately 180K patients. Demographics (age, gender, and ethnicity), diagnosis codes, procedure codes, medication records, lab values, and clinical notes are used as input modalities. The lab measurements includes 111 lab items, such as Hematocrit, Platelet, WBC, Bilirubin, pH, Bicarbonate, Creatinine, Lactate, Potassium, and Sodium. Clinical notes are extracted specifically from the discharge summary section.

\textbf{eICU.} The eICU Collaborative Research Database \cite{pollard2018eicu} is a multi-center critical care dataset that aggregates ICU records from over 200 hospitals in the United States. We use the same six modalities as in MIMIC-IV, except replacing clinical notes with vital signal features. The included modalities are: demographics, diagnoses, procedures, medications, labs, and vitals.

\textbf{Tasks \& Metrics.} For both datasets, we extract data from standardized EHR tables, including \texttt{patients}, \texttt{admissions}, \texttt{prescriptions}, \texttt{diagnoses\_icd}, \texttt{procedures\_icd}, \texttt{labevents}, and \texttt{discharge}. Categorical features such as diagnoses and procedures are encoded using multi-hot medical codes. Lab values are represented using averaged or most recent observations within each admission. ICU visits from patients younger than 18 or older than 89 years, admissions lasting longer than 10 days, and visits that ended in in-hospital mortality are excluded.

We focus on two prediction tasks: (1) Readmission prediction, which aims to predict whether the patient will be readmitted within 15 days following discharge; and (2) Mortality prediction, which predicts whether the patient dies either at discharge (in eICU) or within 90 days post-discharge (in MIMIC-IV). We report both Area Under the Receiver Operating Characteristic Curve (AUC-ROC) and Area Under the Precision-Recall Curve (AUC-PRC) to evaluate model performance.

\textbf{Backbone Encoders.} Following MUSE \cite{wu2024multimodal}, we use Transformer to extract the demographic (age, gender, ethnicity) and sequential medical coding data (diagnosis, procedure, and medication). we use the Recurrent Neural Network to extract time series data (lab values and vital signals) and the TinyBERT to extract text data (clinical notes). All of the modality embeddings are projected to the same latent space with a MLP layer and the dimension is 128.

\subsection{Alzheimer's Disease Dataset}
\textbf{ADNI.} The Alzheimer's Disease Neuroimaging Initiative (ADNI) \cite{jack2008alzheimer} is a longitudinal cohort with over 2K patients, including cognitive test scores, neuroimaging, fluid biomarkers, and genetic information. We use the pre-processed version provided by the TADPOLE challenge \cite{marinescu2019tadpole}, which includes 1,737 patients and 12,741 visits. The multimodal features consist of magnetic resonance imaging, positron emission tomography, and diffusion tensor imaging. 

\textbf{Tasks \& Metrics.} The task is to classify patient visits into one of three diagnostic categories: normal cognition, mild cognitive impairment, and Alzheimer's disease. This is formulated as a \textbf{multi-class classification} problem. We adopt the official TADPOLE metrics: balanced accuracy and the one-vs-one macro AUC-ROC score.

\textbf{Backbone Encoders.} As the TADPOLE challenge gives extracted features for each modality, we use a MLP layer as the backbone encoder and set the dimension to be 128.

\subsection{Atrial Fibrillation Dataset}
\textbf{In-Hospital AFib.}
We use a proprietary dataset of hospitalized atrial fibrillation (AFib) patients with catheter ablation records, collected from UC Davis Medical Center. The dataset contains multimodal modalities: (1) demographic and medical features, (2) intracardiac electrograms (EGM), and (3) metabolic labtest results.

The dataset includes 89 AFib patients in total. 65 patients have available demographic and medical features data with 46 features in total, which include demographics (age and sex), comorbidities (e.g., hypertension, diabetes, CAD, HFpEF), prior treatments (e.g., DOAC, warfarin, pacemaker, ICD), anatomical and physiological measurements (e.g., LVEF, left atrial size, BMI) and ablation-related history (e.g., previous ablation, atypical flutter, CPVA, CTI ablation).

EGM data was collected in the form of raw time series during catheter ablation procedures. The recordings capture bipolar electrograms from the endocardial surface of the heart and were acquired using the CARTO 3D electroanatomical mapping system. A decapolar catheter with 7~mm spacing between bipoles was used to acquire signals at a sampling rate of 1000~Hz. For each patient, electrogram signals were collected both before and after the ablation procedure, covering no fewer than 500 spatial locations. Each recording segment spans approximately 1 to 2 seconds. In total, EGM data are available for 60 patients, representing high-resolution, patient-specific intracardiac electrical activity under clinical ablation settings.

Metabolic laboratory results are available for 43 patients and consist of targeted measurements of inflammatory lipid mediators. These include 46 metabolic markers in range of pro-inflammatory and anti-inflammatory metabolites derived from arachidonic acid, linoleic acid, and omega-3 fatty acids. Specifically, the panel covers epoxy- and hydroxy-fatty acids (e.g., 12(13)-EpODE, 9(10)-EpOME, 11-HETE, 15-HETE), diols (e.g., 9,10-DiHODE, 12,13-DiHOME, 17,18-DiHETE), prostaglandins (e.g., PGF2$\alpha$, 15-deoxy-PGJ2), thromboxanes (e.g., TXB2), and leukotriene pathway products (e.g., 5-HETE, 8-HETE). These metabolites serve as important biomarkers of systemic inflammation, vascular tone, and cardiac remodeling, making them clinically relevant to characterize AFib progression and treatment response.

Since the number of spatial positions and the duration of electrogram recordings vary across patients, we standardize EGM data by uniformly sampling fixed-length time series segments. Specifically, we extracted 1-second window from the raw EGM signals, allowing for overlapping segments. For each patient with available EGM data, we randomly sample 256 positions to construct a single EGM sample representative of that patient. Using this approach, we expand the EGM dataset by a factor of 10, replicating the structured and metabolic modalities of each patient while providing multiple different segments of EGM. The final dataset contains a total of 618 multimodal samples.

Due to real-world clinical constraints, the dataset exhibits natural missing modalities. Not all patients are observed across all three modalities, making this cohort a representative testbed for evaluating multimodal learning under incomplete data conditions. The statistics of the dataset is in Table~\ref{tab:afibstastics}.

\begin{table}[t]
\centering
\caption{AFib Dataset statistics.}
\setlength{\tabcolsep}{8pt}
\renewcommand{\arraystretch}{1}
\begin{tabular}{l p{3.6cm}}
\toprule
\textbf{Item} & \textbf{AFib} \\
\midrule
\#Patients & 89 \\
\#Samples & 618 \\
Classification Labels & Paroxysmal: 0.41,\ Persistent: 0.49,\ Long-persistent: 0.10 \\
Recurrence Labels & Positive: 0.27,\ Negative: 0.73 \\
Age & 72 \\
Gender & F: 0.36,\ M: 0.64 \\
Metabolic Missing (\%) & 0.45 \\
Demographic / Medical Missing \% & 0.17 \\
EGM Missing (\%)  & 0.23 \\
\bottomrule
\end{tabular}
\label{tab:afibstastics}
\end{table}

\textbf{Tasks \& Metrics.} The task is to predict the types of Afib including Normal, Paroxysmal, Persistent, Long-persistent,  and AFib recurrence within 9 months, which is multi-class classification and binary classification tasks, respectively, with evaluation metrics including AUC-ROC, and F1-score.

\textbf{Backbone Encoders.} we use Transformer to extract the demographic and medical features data, the MLP to extract the metabolic data. We use the Recurrent Neural Network to extract EGM data per position and use a 2-layer GCN to project the features from 256 positions to be the learned embedding. All of the modality embeddings are projected to the same latent space with a MLP layer with dimension as 128.

\section{Additional Details on Baselines}
\label{appendix:baselines}
\textbf{CM-AE \cite{ngiam2011multimodal}.}
CM-AE is a cross-modal autoencoder that reconstructs missing modalities from observed ones. It is trained on modality-complete subset to reconstruct missing modalities. We report the results on MIMIC-IV, eICU, and ADNI datasets.

\textbf{SMIL \cite{ma2021smil}.}
SMIL adopts a Bayesian meta-learning approach to estimate the missing modality embeddings in a latent space. It approximates the missing modality using a weighted sum of modality priors, learned from complete cases, combined with a feature-level regularizer.

\textbf{MT \cite{ma2022multimodal}.}
MT applies a transformer-based late-fusion strategy. It uses modality-specific encoders to embed each modality and then fuses them via a transformer layer that learns cross-modal dependencies. We use this as a representative direct prediction baseline. The dual-branch neural network design is achieved by assigning weights to the multi-head attention layers in the two parallel Transformers.

\textbf{GRAPE \cite{you2020handling}.}
GRAPE builds a bipartite graph between patients and modalities, where observed modalities are linked to patients and message passing is used to obtain patient representations. This model implicitly handles missing modalities via graph structure and edge dropout.

\textbf{HGMF \cite{chen2020hgmf}.}
HGMF adopts a heterogeneous graph matching framework, representing patients and modalities as hypernodes and modeling missingness patterns through intra- and inter-node interactions. We use a modified inductive version for fair comparison under partial observations.

\textbf{M3Care \cite{zhang2022m3care}.}
M3Care learns modality-specific patient similarity graphs and aggregates them to form multimodal representations. Each modality builds its own patient graph, and a GNN is applied to integrate across graphs using a Transformer head. The CaD design is integrated with the Transformer head.

\textbf{MUSE \cite{wu2024multimodal}.}
MUSE employs a mutual-consistent contrastive learning strategy over a bipartite graph to address both missing modalities and missing labels. It learns robust patient-modality interactions by enforcing consistency across augmented views, being a enhanced version of GRAPE \cite{you2020handling} considering the relationships between different patients. MUSE+ leverages a semi-supervised learning strategy to explore unlabeled data, which differs from our research focus. Furthermore, based on the principle of independent causal mechanisms \cite{pearl2009causality}, semi-supervised learning is generally ineffective unless there exists a causal relationship from the label $\mathbf{Y}$ to the observed data $\mathbf{X}_O$. We plan to further investigate semi-supervised learning methods from a causal perspective in future work.

\section{Limitations and Future Work}~\label{sec6}
\label{appendix:limitation}

While our framework improves robustness and deconfounds multimodal prediction, it also introduces certain limitations. First, CaD involves additional computational overhead due to the dual-branch neural network and pre-trained backbone to process the confounder dictionary. Although this design improves robustness under missingness, it increases a few training complexity. Second, our approach focuses on debiasing through causal inference techniques, but it does not aim to explicitly uncover mechanistic or physiological disease pathways. The representations learned through our disentanglement may align with causal features but are not guaranteed to reflect clinical mechanisms. Finally, as noted in Remark~1, the theoretical validity of our causal objective requires that the data contain sufficient information to block all confounding paths. In real-world applications, this condition may not always hold due to data acquisition conditions. Hence, the integration of an explicit causal structure from multimodal data with the proposed model would be the subject of future work. 

\bibliographystyle{IEEEtran}
\bibliography{IEEEabrv,ref}